\documentclass[journal]{IEEEtran}
\ifCLASSINFOpdf
\else
\fi

\usepackage{amssymb}
\usepackage{url}
\usepackage{algorithm} 
\usepackage{algorithmic} 

\usepackage{subfigure}
\usepackage{amsmath} 
\usepackage{graphicx}
\usepackage{multirow}

\begin{document}
%
\title{Learning idempotent representation for subspace clustering}
%
%
%

\author{Lai~Wei,
        Shiteng Liu,
        Rigui~Zhou and Changming~Zhu 
\thanks{L. Wei, S. Liu, R. Zhou, C. Zhu were with the College of Information Engineering, Shanghai Maritime University, Shanghai, 201306 China (e-mail: weilai@shmtu.edu.cn, Liushiteng1218@163.com, rgzhou@shmtu.edu.cn, cmzhu@shmtu.edu.cn.)}}

%
%

\markboth{Submitted to IEEE transactions on Knowledge and Data Engineering}%
{Shell \MakeLowercase{\textit{et al.}}: Bare Demo of IEEEtran.cls for IEEE Journals}
%



\maketitle

\begin{abstract}
The critical point for the successes of spectral-type subspace clustering algorithms is to seek reconstruction coefficient matrices which can faithfully reveal the subspace structures of data sets. An ideal reconstruction coefficient matrix should have two properties: 1) it is block diagonal with each block indicating a subspace; 2) each block is fully connected. Though there are various spectral-type subspace clustering algorithms have been proposed, some defects still exist in the reconstruction coefficient matrices constructed by these algorithms. We find that a normalized membership matrix naturally satisfies the above two conditions. Therefore, in this paper, we devise an idempotent representation (IDR) algorithm to pursue reconstruction coefficient matrices approximating normalized membership matrices. IDR designs a new idempotent constraint for reconstruction coefficient matrices. And by combining the doubly stochastic constraints, the coefficient matrices which are closed to normalized membership matrices could be directly achieved. We present the optimization algorithm for solving IDR problem and analyze its computation burden as well as convergence. The comparisons between IDR and related algorithms show the superiority of IDR. Plentiful experiments conducted on both synthetic and real world datasets prove that IDR is an effective and efficient subspace clustering algorithm.
\end{abstract}

\begin{IEEEkeywords}
subspace clustering, idempotent matrix, doubly stochastic constraint, normalized membership matrix
\end{IEEEkeywords}

%
\IEEEpeerreviewmaketitle

\section{Introduction}
\label{sec1}
%
%
%
%
\IEEEPARstart{H}{igh}-dimensional data samples emerged in computer vision fields could be viewed as generated from a union of linear subspaces \cite{art_1,RN2518,DBLP:conf/sigmod/AgrawalGGR98,DBLP:journals/tkdd/KriegelKZ09}. subspace clustering, whose goal is to partition the data samples into several clusters with each cluster corresponding to a subspace, has attracted lots of researchers' attentions. In the past decades, many kinds of subspace clustering algorithms have been proposed \cite{RN1940,DBLP:journals/tkde/ChuCYC09,DBLP:journals/tcsv/YiHCC18,DBLP:journals/pami/MaDHW07,DBLP:journals/pami/ElhamifarV13,RN1710}. Among them, spectral-type methods showed more excellent performances in many applications such as motion segmentation, face clustering and so on \cite{DBLP:journals/pami/ElhamifarV13,RN1710,DBLP:conf/iccv/Kanatani01,DBLP:journals/siamrev/MaYDF08}. 
\par Without loss of generality, suppose that a clean data matrix $\mathbf{X}= [\mathbf{X}_1,\mathbf{X}_2,$ $\cdots,\mathbf{X}_k]\in\mathcal{R}^{d\times n}$ contains $n$ data samples drawn from $k$ subspaces. $\mathbf{X}_i\subset\mathbf{X}$ denotes the sub-matrix including $n_i$ data samples lying in the $i$-th subspace, where $\sum_{i=1}^hn_i=n$. And if $i\neq j$ ($i,j =1,2,\cdots,k$), $\mathbf{X}_i\cap\mathbf{X}_j=\emptyset$. The framework of spectral-type subspace clustering algorithms is divided into three parts. Firstly, they learn a reconstruction coefficient matrix $\mathbf{Z}\in \mathcal{R}^{n\times n}$ satisfying $\mathbf{X}=\mathbf{XZ}$. Secondly, an affinity matrix $\mathbf{A}$ is built by using the obtained reconstruction coefficient matrix, i.e. $[\mathbf{A}]_{ij}=(|[\mathbf{Z}]_{ij}| + |[\mathbf{Z}^{\top}]_{ij}|)/2$, where $[\mathbf{A}]_{ij}$ and $[\mathbf{Z}]_{ij}$ denote the $(i,j)-th$ element of $\mathbf{A}$ and $\mathbf{Z}$ respectively, $\mathbf{Z}^{\top}$ is the transpose of $\mathbf{Z}$. Finally, a certain spectral clustering algorithm, e.g. normalized cuts (Ncuts) \cite{DBLP:journals/pami/ShiM00}, is used to get the final clustering results by using $\mathbf{A}$. It could be clearly seen that the performance of a spectral-type algorithm mainly rely on the learned reconstruction matrix. An ideal coefficient matrix should have inter-subspace sparsity and intra-subspace connectivity. Namely, if $\mathbf{x}_i$ and $\mathbf{x}_j$ belong to a same subspace, $|[\mathbf{Z}]_{ij}|>0$. Otherwise, $|[\mathbf{Z}]_{ij}|=0$. 
\par Different spectral-type methods use different regularizers to produce coefficient matrices with different characteristics. For instance, sparse subspace clustering (SSC) \cite{DBLP:journals/pami/ElhamifarV13,conf_1} pursuits sparse reconstruction coefficient matrices by introducing a sparse constraint \cite{RN802}. Low-rank representation (LRR) \cite{RN1710,conf_2} seeks a low-rank reconstruct coefficient matrix by minimize the nuclear norm of the coefficient matrix. Least square regression (LSR) \cite{Lu:2012} defines a Frobenuis norm regularizer and searches a dense reconstruction coefficient matrix. Block diagonal representation (BDR) \cite{RN2485} provides a $k$ block diagonal reconstruction coefficient matrix by minimizing the sum of smallest $k$ eigenvalues of the coefficient matrix's Laplacian regularizer. Though these representative methods achieve promising results in different kinds of subspace clustering tasks, the obtained coefficient matrices still have some drawbacks. The coefficient matrices gotten by SSC are usually too sparse to lack connectedness within each subspace. The block diagonal constraint used in BDR may not lead the correct clustering, since each block still may not be fully connected. On the other hand, although the connectedness within subspaces are guaranteed in the dense coefficient matrices constructed by LRR and LSR, the coefficients of the inter-subspace samples are usually non-zero. In order to get away the dilemmas, three different types of methods are emerged. Firstly, some extensions of classical regularizers are developed. For example, Zhang et al. extended the nuclear norm regularizer used in LRR to a kind of Schatten-$p$ norm regularizer \cite{RN2478}. Xu et al. proposed a scaled simplex representation by adding the non-negative constraint and scaled affine constraint of the coefficient matrix obtained in LSR \cite{RN2691}. Secondly, researchers begin to use mixed regularizers of coefficient matrices. Li et al. proposed a structured sparse subspace clustering (SSSC) \cite{RN2328} by adding a re-weighted $l_1$-norm regularizer into SSC. Elastic net (EN) method defined a combination of $l_1$-norm and Frobenius regularizer of the coefficient matrices \cite{RN1705,RN2562}. Zhuang et al. combined sparse constraint and nuclear norm regularizer together to propose a non-negative low-rank and sparse representation method (NNLRSR) \cite{DBLP:journals/tip/ZhuangGTWLMY15}. Tang et al. generalized NNLRSR and devised a structure-constrained LRR (SCLRR) \cite{RN1888}. Lu et al. presented a graph-regularized LRR (GLRR) algorithm by minimizing the nuclear norm and the Laplacian regularizer the coefficient matrix simultaneously \cite{RN1946}. Tang et al. designed a dense block and sparse representation (DBSR) method which used the $2$-norm (the maximal singular value) and $l_1$ norm regularizers to compute a dense block and sparse coefficient matrix \cite{RN2192}. Thirdly, classical spectral-type subspace clustering algorithms are integrated to build cascade models. Wei et al. devised a sparse relation representation by stacking SSC and LRR \cite{RN2723}. Sui et al. also provided a similar method to show the effectiveness of cascade models \cite{RN2532}. These extended methods outperform the classical algorithms to a certain extent, but they still may not guarantee to produce ideal coefficient matrices. 
\par From the view point of a spectral clustering algorithm, the best affinity matrix $\mathbf{M}^*$ of data set $\mathbf{X}$ should have the following properties: If $\mathbf{x}_i$ and $\mathbf{x}_j$ belong to a same cluster, then $[\mathbf{M}^*]_{ij} = 1$. Otherwise, $[\mathbf{M}^*]_{ij} = 0$. Namely, $\mathbf{M}^*$ should be $k$ block diagonal and have the following formulation:
\begin{equation}\label{e1}
  \mathbf{M}=\left(
    \begin{array}{cccc}
      \mathbf{1}_{n_1}\mathbf{1}_{n_1}^{\top}& 0 & \cdots & 0 \\
      0 & \mathbf{1}_{n_2}\mathbf{1}^{\top}_{n_2} & \cdots & 0 \\
      \vdots  & \ddots & \vdots & \vdots \\
      0 & \cdots & 0 & \mathbf{1}_{n_k}\mathbf{1}^{\top}_{n_k}
     \end{array}
  \right) 
\end{equation}
where $\mathbf{1}_{n_i}$ is a column vector with $n_i$ elements and each element equals $1$.  In the correlation clustering domain, $\mathbf{M}^*$ is called a \textbf{membership matrix} \cite{DBLP:conf/soda/MathieuS10,DBLP:conf/soda/Swamy04,DBLP:conf/cvpr/LeeLLK15}. Moreover, the researchers also proved that a variation of the membership matrix, called \textbf{normalized membership matrix}, is also adequate for spectral clustering. The normalized membership matrix $\mathbf{A}^*$ corresponding to $\mathbf{M}^*$ is expressed as follows:
\begin{equation}\label{e2}
  \mathbf{A}^*=\left(
    \begin{array}{cccc}
      \frac{1}{n_1}\mathbf{1}_{n_1}\mathbf{1}_{n_1}^{\top}& 0 & \cdots & 0 \\
      0 & \frac{1}{n_2}\mathbf{1}_{n_2}\mathbf{1}^{\top}_{n_2} & \cdots & 0 \\
      \vdots  & \ddots & \vdots & \vdots \\
      0 & \cdots & 0 & \frac{1}{n_k}\mathbf{1}_{n_k}\mathbf{1}^{\top}_{n_k}
     \end{array}
  \right). 
\end{equation}
\par
\begin{figure}
  \centering
  \includegraphics[width=0.48\textwidth]{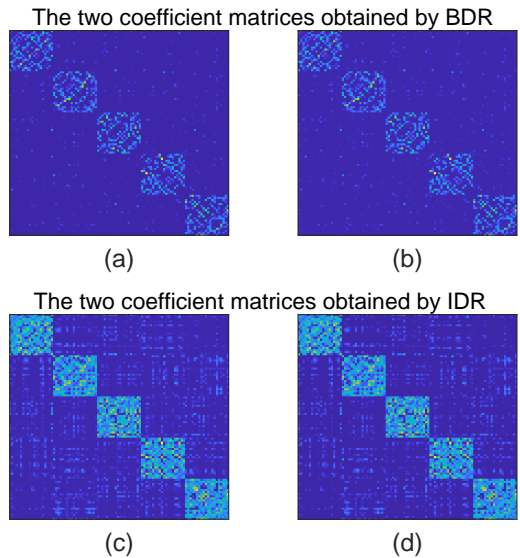}\\
  \caption{We generate $5$ subspaces each of dimension $d = 5$ in an ambient space of dimension $d = 20$. We sampled $50$ data points from each subspace and constructed a $d\times 250$ data matrix $\mathbf{X}=[\mathbf{X}_1,\cdots,\mathbf{X}_5]$ without noise. $\mathbf{X}_i (i=1,2,\cdots,5)$ contains the samples from $i-th$ subspace. Then we use BDR and the proposed algorithm in this paper to compute the coefficient matrices. The two coefficient matrices obtained by BDR are illustrated in (a) and (b). The two coefficient matrices achieved by the proposed algorithm are shown in (c) and (d). }\label{f1}
\end{figure}

 Back to the domain of subspace clustering, suppose an affinity matrix $\mathbf{A}$ is a normalized membership matrix. As we mentioned above, in a spectral-type subspace clustering algorithm, an affinity matrix is defined as $[\mathbf{A}]_{ij}=(|[\mathbf{Z}]_{ij}| + |[\mathbf{Z}^{\top}]_{ij}|)/2$. If we force $\mathbf{Z}=\mathbf{Z}^{\top}$ and $[\mathbf{Z}]_{ij}\geq 0$ (for all $i,j$), the best reconstruction coefficient matrix $\mathbf{Z}^*$ should be same as $\mathbf{A}^*$, namely $\mathbf{Z}^*$ is also a normalized membership matrix. We can see such $\mathbf{Z}^*$ is definitely inter-subspace sparse and intra-subspace connective. The property that each element in a block (i.e., $\mathbf{Z}_i^*$) equals $1/n_i(>0)$ means $\mathbf{Z}^*$ is fully connected in each block. Hence, this kind of coefficient matrix is better than that obtained by BDR. Fig. \ref{f1} presents the two coefficient matrices obtained by BDR and the proposed algorithm in this paper on a synthetic data set. We can see that all coefficient matrices are block diagonal, but each block in the coefficient matrices obtained by the proposed algorithms is denser than that obtained by BDR. Then the subspace structure of the data set could be revealed more faithfully by using the proposed algorithm. 

\par In \cite{DBLP:conf/cvpr/LeeLLK15}, Lee et al. also suggested constructing a normalized membership matrix for subspace clustering. However, the so-called membership representation (MR) algorithm \cite{DBLP:conf/cvpr/LeeLLK15} takes three steps to finally get the coefficient matrix. Firstly, a certain subspace clustering algorithm, such as SSC or LRR, was used to get an initial coefficient matrix. Secondly, MR sought a membership matrix by using the obtained initial coefficient matrix. Finally, a normalized membership matrix was computed with the obtained membership matrix. In the last two steps, augmented Lagrangian method (ALM) \cite{DBLP:journals/corr/LinCM10} was applied to solve the corresponding optimization problems. Hence, besides the computation time used for finding an initial coefficient matrix, the time cost in the last two steps of MR was also high. 
\par In this paper, we invent a new method to find a coefficient matrix which is closed to a normalized membership matrix as much as possible. The motivation of the proposed algorithm is the \textbf{self-expressiveness property of the reconstruction coefficient vectors} obtained by subspace clustering algorithms. As we know, spectral-type subspace clustering algorithms assume that the original data samples obey the self-expressiveness property \cite{DBLP:journals/pami/ElhamifarV13}, i.e., each data point can be well reconstructed by a linear combination of other points in the given dataset. The self-expressiveness property of the obtained coefficient vectors means each coefficient vector could be linearly reconstructed by other coefficient vectors. Based on this proposition and the doubly stochastic constraints \cite{DBLP:conf/iccv/ZassS05,DBLP:conf/nips/ZassS06a}, an idempotent representation (IDR) method for subspace clustering is proposed. For solving the IDR problem, an optimization algorithm is also presented. And the convergence as well as the complexity analysis of the optimization algorithm are given consequently. We also make comparisons between IDR and some related algorithms, so that the superiority of IDR is shown. Finally, extensive experiments conducted on both synthetic and real world databases show the effectiveness and efficiency of IDR method.
\par The rest of the paper is organized as follows: we introduce the general formulation of spectral-type subspace clustering algorithms in Section \ref{sec2}. In Section \ref{sec3}, we propose the idea of idempotent representation (IDR) and the optimization algorithm for solving IDR problem. The further discussions of IDR, such as the analysis on the convergence and complexity of the optimization algorithm, the connections between IDR and the related algorithms, are given in Section \ref{sec4}. Comparative subspace clustering experiments on both synthetic data set and real world data sets are performed in Section \ref{sec5}. Section \ref{sec6} presents the conclusions.
\section{Preliminary}
\label{sec2}
Through there is a wide variety of existing spectral-type subspace clustering algorithms, the general objective function of these algorithms could be expressed as follows:
\begin{equation}\label{e3}
  \begin{array}{ll}
    \min_{\mathbf{Z,E}} & \Omega (\mathbf{Z})  \\
    s.t.                & \mathbf{X} = \mathbf{XZ}, 
  \end{array}
\end{equation}
where $\Omega (\mathbf{Z})$ indicates a certain norm regularizer of $\mathbf{Z}$, $\mathbf{X}\in \mathcal{R}^{d\times n}$ is a data matrix. In real applications, data is often noisy or corrputed. Hence, the more robust version of above problem could be defined as follows:
\begin{equation}\label{e4}
  \begin{array}{ll}
    \min_{\mathbf{Z,E}} & \Omega (\mathbf{Z}) + \lambda \Phi (\mathbf{E}), \\
    s.t.                & \mathbf{X} = \mathbf{XZ} + \mathbf{E}, 
  \end{array}
\end{equation}
where $\mathbf{E}$ is the error term, $\Phi(\mathbf{E})$ is a certain measurement of $\mathbf{E}$. $\lambda$ is a positive parameter which is used to balance the effects of $ \Omega(\mathbf{Z})$ and $\Phi(\mathbf{E})$. Moreover, some algorithms add some additional constraints of $\mathbf{Z}$ which could be expressed as $\Theta (\mathbf{Z})$. Then the main differences between the existing subspace clustering algorithms are the definitions of $\Omega(\cdot),\Phi(\cdot)$ and $\Theta(\cdot)$. We use Table \ref{t1} to summarize the formulations of $\Omega(\mathbf{Z}),\Phi(\mathbf{E})$ and $\Theta(\mathbf{Z})$ of some representative subspace clustering algorithms.
\begin{table}
  \begin{center}
    \scriptsize 
  \caption{The residual terms, regularizers and additional constraints of coefficient matrices used in some subspace clustering algorithms.}\label{t1}
  
  \begin{tabular}{l|l|l|l}
    \hline
   Algorithms     & $\Omega(\mathbf{Z})$       & $\Phi(\mathbf{E})$         & $\Theta(\mathbf{Z})$\\ \hline
   SSC            & $\|\mathbf{Z}\|_1$    & $\|\mathbf{E}\|_1$      & $diag(\mathbf{Z})=\mathbf{0}_n$  \\
   LRR            & $\|\mathbf{Z}\|_*$    & $\|\mathbf{E}\|_{2,1}$  & -\\
   LSR            & $\|\mathbf{Z}\|_F^2$  & $\|\mathbf{E}\|_F^2$    & $diag(\mathbf{Z})=\mathbf{0}_n$  \\
   BDR            & $\|\mathbf{Z}\|_{k}$    & $\|\mathbf{E}\|_F^2$    & $diag(\mathbf{Z})=\mathbf{0}_n,$  \\
   &                       &                         & $\mathbf{Z}=\mathbf{Z}^{\top},$ \\
   &                       &                         & $\mathbf{Z}\geq 0$ \\
   SSSC    & $\|(\mathbf{I}_n + \gamma\mathbf{Q})\bigodot \mathbf{Z}\|_1$    & $\|\mathbf{E}\|_1$      & $diag(\mathbf{Z})=\mathbf{0}_n$\\
   EN          & $\|\mathbf{Z}\|_F^2$ +$\gamma\|\mathbf{Z}\|_1$    & $\|\mathbf{E}\|_{1}$      & $diag(\mathbf{Z})=\mathbf{0}_n$\\ 
   SCLRR          & $\|\mathbf{Z}\|_*$ +$\gamma\|\mathbf{Z}\|_1$    & $\|\mathbf{E}\|_{2,1}$      & -\\
   GLRR           & $\|\mathbf{Z}\|_*$ + $\gamma Tr(\mathbf{ZLZ}^{\top})$      & $\|\mathbf{E}\|_{2,1}$      & -\\
   DBSR           & $\|\mathbf{Z}\|_2$ + $\gamma\|\mathbf{Z}\|_1$      & $\|\mathbf{E}\|_{2,1}$      & -\\
   \hline
  \end{tabular}
  \end{center}
  Notice: $\gamma>0$ is a parameter. $diag(\mathbf{Z})$ denotes a column vector composed by the elements in the diagonal of $\mathbf{Z}$. $\mathbf{0}_n$ is a column vector with each element equals $0$. In BDR, $\|\mathbf{Z}\|_{k}$ is a constraint which forces $\mathbf{Z}$ to be $k$ block diagonal. In SSSC, $\mathbf{Q}\in \mathcal{R}^{n\times n}$ is a weighted matrix updated by the segmentation results in each iteration and $\mathbf{I}_n$ is an $n\times n$ identity matrix. In GLRR, $Tr(\cdot)$ denotes the trace of a matrix and $\mathbf{L}$ is the Laplacian matrix built by using K-nearest-neighbors (KNN) \cite{DBLP:books/lib/DudaHS01} and $\mathbf{X}$. 
\end{table}
\par All the algorithms mentioned in Table \ref{t1} could be solved by using ALM. Consequently, Ncuts is used to get final subspace clustering results.

\section{Idempotent representation}
\label{sec3}
\subsection{Motivation}
\label{sec3.1}
The key point of the spectral-type subspace clustering algorithms have in common is that they all assume that the data samples in $\mathbf{X}$ obey the self-expressiveness property \cite{DBLP:journals/pami/ElhamifarV13}. Namely, each data sample could be approximately constructed by a linear combination of other data points in the given dataset with tolerable errors. Thus, $\mathbf{X}\approx \mathbf{XZ}$ and $\mathbf{Z}$ records the reconstruction relationship of the original data samples.
\par In addition, as described in \cite{RN1710,conf_2}, the obtained coefficient matrix $\mathbf{Z}$ is a representation of the original data matrix $\mathbf{X}$ with $\mathbf{z}_i$ being the representation of $\mathbf{x}_i$. Here, $\mathbf{z}_i$ and $\mathbf{x}_i$ are the $i$-th columns of $\mathbf{Z}$ and $\mathbf{X}$ respectively. Then it is reasonable to assume that the coefficient vectors also obey the self-expressiveness property (\textbf{ Self-expressiveness property of coefficient vectors}), namely each coefficient vector could be linearly reconstructed by other coefficient vectors in $\mathbf{Z}$. Therefore, we have 
\begin{equation}\label{en1}
  \mathbf{Z}=\mathbf{ZT},
\end{equation}
where $\mathbf{T}$ is a reconstruction coefficient matrix corresponding to $\mathbf{Z}$. Moreover, we could hope $\mathbf{T}$ to be closed to $\mathbf{Z}$. Because if $\mathbf{Z}$ is a good representation of $\mathbf{X}$, $\mathbf{Z}$ should follow the reconstruction relationship of the original data set and $\mathbf{Z}$ records the reconstruction relationship of the original data samples. Therefore, the following equation holds 
\begin{equation}\label{e5}
\mathbf{Z}\approx \mathbf{Z}\times\mathbf{Z} = \mathbf{Z}^2.
\end{equation}
The above equation means that $\mathbf{Z}$ is approximate to an idempotent matrix.
\par It is easy to verify that an $n\times n$ identity matrix $\mathbf{I}_n$ is idempotent and the solution to the problem $\mathbf{X}=\mathbf{XZ}$. Then in sepctral-type subspace clustering algorithm, the above idempotent constraint (Eq. (\ref{e5})) is not sufficient for finding a good coefficient matrix. Fortunately, it could be checked that a normalized membership matrix is also an idempotent matrix. Hence, we will show how to add some necessary constraints to compel an idempotent reconstruction coefficient matrix to be a normalized membership matrix. 
\par In fact, Lee et al. pointed that an idempotent matrix is a normalized membership matrix if and only if it is doubly stochastic \cite{DBLP:conf/cvpr/LeeLLK15}. And a doubly stochastic matrix $\mathbf{Z}\in \mathcal{R}^{n\times n}$ can be completely described by the following doubly stochastic conditions \cite{DBLP:conf/iccv/ZassS05,DBLP:conf/nips/ZassS06a}:
\begin{equation}\label{e6}
   \mathbf{1}_n^{\top}\mathbf{Z} = \mathbf{1}_n^{\top}, \mathbf{Z} = \mathbf{Z}^{\top}, \mathbf{Z} \geq \mathbf{0}.
\end{equation}
However, these conditions still can not prevent $\mathbf{Z}$ to be $\mathbf{I}_n$. As mentioned above, for revealing the subspace structure of data set $\mathbf{X}$ with $k$ subspaces faithfully, a coefficient matrix should be $k$ block diagonal. Then for an idempotent and doubly stochastic coefficient matrix $\mathbf{Z}$, we could simply let $Tr(\mathbf{Z})=k$, then $\mathbf{Z}$ would be $k$ block diagonal. This constraint could also prevent $\mathbf{Z}$ to degenerate the trivial solution, i.e., $\mathbf{Z}=\mathbf{I}_n$. Therefore, by integrating these constraints and the general formulation of subspace clustering algorithms, we could define the idempotent representation (IDR) problem as follows:
\begin{equation}\label{e7}
  \begin{array}{ll}
    \min_{\mathbf{Z}} & \|\mathbf{Z}\|_{id} + \lambda \|\mathbf{E}\|_{2,1}, \\
    s.t.                & \mathbf{X} = \mathbf{XZ} + \mathbf{E},\\
    & \mathbf{1}_n^{\top}\mathbf{Z} = \mathbf{1}_n^{\top}, \mathbf{Z} = \mathbf{Z}^{\top},\mathbf{Z} \geq \mathbf{0},Tr(\mathbf{Z})=k, 
  \end{array}
\end{equation}
where $\|\mathbf{Z}\|_{id}$ denotes an idempotent regularizer of $\mathbf{Z}$, namely $\|\mathbf{Z}\|_{id} = \|\mathbf{Z}-\mathbf{Z}^2\|_F^2$. In most real applications, partial data samples are corrupted, hence we use $l_{2,1}$ norm to measure the error term $\mathbf{E}$.
\par Furthermore, all these restrictions imposed on $\mathbf{Z}$ will limit its representation capability. For alleviating this problem, we introduce an intermediate term and proposed the following relaxed problem:
\begin{equation}\label{e8}
  \begin{array}{ll}
    \min_{\mathbf{Z,S}} & \|\mathbf{Z}-\mathbf{S}\|_F^2+\gamma\|\mathbf{S}\|_{id} + \lambda \|\mathbf{E}\|_{2,1}, \\
    s.t.            &  \mathbf{X} = \mathbf{XZ} + \mathbf{E},   \\
    & \mathbf{1}_n^{\top}\mathbf{S} = \mathbf{1}_n^{\top}, \mathbf{S} = \mathbf{S}^{\top},\mathbf{S} \geq \mathbf{0},Tr(\mathbf{S})=k,
  \end{array}
\end{equation}
where $\gamma$ is also a positive parameter. 
\subsection{Optimization}
Similar to solving the existing subspace clustering problems, we use ALM \cite{DBLP:journals/corr/LinCM10} to find the solutions to IDR problem (e.g., Eq. (\ref{e8})). Firstly, we need to transfer Eq. (\ref{e8}) to the following equivalent problem:
\begin{equation}\label{e9}
  \begin{array}{ll}
    \min_{\mathbf{Z,S,C,D}} & \|\mathbf{Z}-\mathbf{S}\|_F^2+\gamma\|\mathbf{S}-\mathbf{SC}\|_F^2 + \lambda \|\mathbf{E}\|_{2,1}, \\
    s.t. &  \mathbf{X} = \mathbf{XZ} + \mathbf{E},\\                
    & \mathbf{S} = \mathbf{C},\mathbf{S} = \mathbf{S}^{\top},\mathbf{S} \geq \mathbf{0},\\
    &\mathbf{1}_n^{\top}\mathbf{C} = \mathbf{1}_n^{\top},\\
                        & \mathbf{S} = \mathbf{D},Tr(\mathbf{D})=k,
  \end{array}
\end{equation}
where $\mathbf{C},\mathbf{D}$ are two auxiliary variables. Then the corresponding augmented Lagrangian function of Eq. (\ref{e9}) could be expressed as follows:
\begin{equation}\label{e10}
  \begin{array}{ll}
    \mathfrak{L} & = \|\mathbf{Z}-\mathbf{S}\|_F^2+\gamma\|\mathbf{S}-\mathbf{SC}\|_F^2 + \lambda \|\mathbf{E}\|_{2,1} \\
    &+<\mathbf{Y}_1, \mathbf{X} - \mathbf{XZ} - \mathbf{E}> + <\mathbf{Y}_2,\mathbf{S} - \mathbf{C}>\\
    &+ <\mathbf{Y}_3,\mathbf{1}_n^{\top}\mathbf{C} - \mathbf{1}_n^{\top}>+<\mathbf{Y}_4,\mathbf{S} - \mathbf{D}>\\
    &+ \mu/2\big(\|\mathbf{X} - \mathbf{XZ} - \mathbf{E}\|_F^2+\|\mathbf{S} - \mathbf{C}\|_F^2\\
    &+\|\mathbf{1}_n^{\top}\mathbf{C} - \mathbf{1}_n^{\top}\|_F^2+\|\mathbf{S} - \mathbf{D}\|_F^2\big),
  \end{array}
\end{equation}
where $\mathbf{Y}_1,\mathbf{Y}_2$, $\mathbf{Y}_3$ and $\mathbf{Y}_4$ are four Langrangian multipliers and $\mu>0$ is an additional parameter. By minimizing $\mathcal{L}$, the variables ${\mathbf{Z},\mathbf{S},\mathbf{C},\mathbf{D},\mathbf{E}}$ could be optimized alternately while fixing others.
\par \textbf{1. Fix other variables and update $\mathbf{Z}$.} Then in the $h-th$ iteration ($h$ is the number of iterations), 
\begin{equation}\label{e11}
  \begin{array}{l}
  \mathbf{Z}^{h+1} = \arg \min_{\mathbf{Z}}\|\mathbf{Z}-\mathbf{S}^h\|_F^2 + <\mathbf{Y}_1^h,\mathbf{X} - \mathbf{XZ} - \mathbf{E}^{h}>\\
  +\mu^h/2\|\mathbf{X} - \mathbf{XZ} - \mathbf{E}^{h}\|_F^2\\
  =\arg \min_{\mathbf{Z}}\|\mathbf{Z}-\mathbf{S}^h\|_F^2 + \mu^h/2\|\mathbf{X} - \mathbf{XZ} - \mathbf{E}^{h} + \mathbf{Y}^h_1/\mu^h\|_F^2,
  \end{array}
\end{equation}
where $\mathbf{S}^h,\mathbf{Y}^h_1$ and $\mu^h$ are the updated variables. It could be easily verified that $\mathbf{Z}^{h+1} = \big(2\mathbf{I}_n + \mu^h\mathbf{X}^{\top}\mathbf{X}\big)^{-1}\big(2\mathbf{S}^h + \mu^h(\mathbf{X}^{\top}\mathbf{X} - \mathbf{X}^{\top}\mathbf{E}^h) + \mathbf{X}^{\top}\mathbf{Y}_1^h\big)$, where $\big(2\mathbf{I}_n + \mu^h\mathbf{X}^{\top}\mathbf{X}\big)^{-1}$ is the pseudo-inverse of $(2\mathbf{I}_n + \mu^h\mathbf{X}^{\top}\mathbf{X})$.
\par \textbf{2. Fix other variables and update $\mathbf{S}$.} Similar to updting $\mathbf{Z}$, 
\begin{equation}\label{e12}
  \begin{array}{l}
  \mathbf{S}^{h+1} = \arg \min_{\mathbf{S}}\|\mathbf{Z}^{h+1}-\mathbf{S}\|_F^2 + \gamma\|\mathbf{S}-\mathbf{SC}^h\|_F^2
  \\+<\mathbf{Y}^h_2,\mathbf{S}-\mathbf{C}^h> +<\mathbf{Y}^h_4,\mathbf{S}-\mathbf{D}^h>
  \\+ \mu^h/2\big(\|\mathbf{S}-\mathbf{C}^h\|_F^2+\|\mathbf{S}-\mathbf{D}^h\|_F^2\big)
  \\= \arg \min_{\mathbf{S}}\|\mathbf{Z}^{h+1}-\mathbf{S}\|_F^2 + \gamma\|\mathbf{S}-\mathbf{SC}^h\|_F^2
  \\+
  \mu^h/2\big(\|\mathbf{S}-\mathbf{C}^h
  +\mathbf{Y}^h_2/\mu^h\|_F^2+\|\mathbf{S}-\mathbf{D}^h+\mathbf{Y}^h_4/\mu^h\|_F^2\big).
  \end{array}
\end{equation}
Hence, $\mathbf{S}^{h+1}  = \big(2\mathbf{Z}^{h+1}+\mu^h\mathbf{C}^h-\mathbf{Y}^h_2+\mu^h\mathbf{D}^h-\mathbf{Y}^h_4\big)\big((2+2\mu^h)\mathbf{I}_n+2\gamma(\mathbf{I}_n-\mathbf{C}^h)(\mathbf{I}_n-\mathbf{C}^h)^{\top}\big)^{-1}$.  Because of the non-negative and symmetric constraints on $\mathbf{S}$, we further let $\mathbf{S}^{h+1} = \max (\mathbf{S}^{h+1},\mathbf{0})$ and $\mathbf{S}^{h+1} = \big(\mathbf{S}^{h+1} + (\mathbf{S}^{h+1})^{\top}\big)/2$.
\par \textbf{3. Fix other variables and update $\mathbf{C}$.} We also could find
\begin{equation}\label{e13}
  \begin{array}{l}
  \mathbf{C}^{h+1} = \arg \min_{\mathbf{C}} \gamma\|\mathbf{S}^{h+1}-\mathbf{S}^{h+1}\mathbf{C}\|_F^2+<\mathbf{Y}^h_2,\mathbf{S}^{h+1}-\mathbf{C}>\\
  +<\mathbf{Y}^h_3,\mathbf{1}_n^{\top}\mathbf{C}-\mathbf{1}_n^{\top}>+\mu^h/2\big(\|\mathbf{S}^{h+1}-\mathbf{C}\|_F^2\\
  +\|\mathbf{1}_n^{\top}\mathbf{C}-\mathbf{1}_n^{\top}\|_F^2\big)\\
  = \arg \min_{\mathbf{C}} \gamma\|\mathbf{S}^{h+1}-\mathbf{S}^{h+1}\mathbf{C}\|_F^2 +\mu^h/2\big(\|\mathbf{S}^{h+1}-\mathbf{C}\\
  +\mathbf{Y}^h_2/\mu^h\|_F^2 +\|\mathbf{1}_n^{\top}\mathbf{C} -\mathbf{1}_n^{\top} + \mathbf{Y}^h_3/\mu^h\|_F^2\big).
  \end{array}
\end{equation}
Then $\mathbf{C}^{h+1} = \big(2\gamma(\mathbf{S}^{h+1})^{\top}\mathbf{S}^{h+1}+\mu^h(\mathbf{I}_n + \mathbf{1}_n\mathbf{1}_n^{\top})\big)^{-1}\big(2\gamma(\mathbf{S}^{h+1})^{\top}\mathbf{S}^{h+1}+\mathbf{Y}_2^h -\mathbf{1}_n\mathbf{Y}_3^h+\mu^h(\mathbf{S}^{h+1}+\mathbf{1}_n\mathbf{1}_n^{\top})\big)$.
\par \textbf{4. Fix other variables and update $\mathbf{D}$.} For updating $\mathbf{D}$, we could get the following problem:
\begin{equation}\label{e14}
  \begin{array}{ll}
    \min_{\mathbf{D}} &<\mathbf{Y}_4^h,\mathbf{S}^{h+1}-\mathbf{D}>+\mu^h/2\|\mathbf{S}^{h+1}-\mathbf{D}\|_F^2\\
   =\min_{\mathbf{D}} &\|\mathbf{S}^{h+1}-\mathbf{D}+\mathbf{Y}_4^h/\mu^h\|_F^2\\
   =\min_{\mathbf{D}} &\|\mathbf{D} - \mathbf{M}\|_F^2,\\
   s.t. & Tr(\mathbf{D})=k,
  \end{array}
\end{equation}
where $\mathbf{M} = \mathbf{S}^{h+1}+\mathbf{Y}_4^h/\mu^h$. Note that the constraint is just imposed on the diagonal elements of $\mathbf{D}$, hence $[\mathbf{D}^{h+1}]_{ij} = [\mathbf{M}]_{ij}$, if $i\neq j (i,j=1,2,\cdots,n)$. Let $\mathbf{d} = diag(\mathbf{D})$ and $\mathbf{m} = diag(\mathbf{M})$, then we have 
\begin{equation}\label{e15}
  \begin{array}{ll}
   \min_{\mathbf{d}} &\|\mathbf{d} - \mathbf{m}\|_2^2,\\
   s.t. & \mathbf{1}_n^{\top}\mathbf{d}=k.
  \end{array}
\end{equation}
This problem could be solved by any off-the-shelf quadratic programming solver. We here provide a more efficient method to achieve the solution to Problem (\ref{e15}). The Lagrangian function of Problem (\ref{e15}) is 
\begin{equation}\label{e16}
  \mathcal{L} =\|\mathbf{d} - \mathbf{m}\|_2^2 - \eta(\mathbf{1}_n^{\top}\mathbf{d}-k),
\end{equation}
where $\eta>0$ is a Lagrange multiplier. The optimal solution $\mathbf{d}$ should satisfy that the derivative of Eq. (\ref{e15}) w.r.t. $\mathbf{d}$ is equal to zero, so we have
\begin{equation}\label{e17}
  2(\mathbf{d} - \mathbf{m})-\eta\mathbf{1}_n=\mathbf{0}_n.
\end{equation}
Then for the $j-th$ element of $\mathbf{d}$, we have
\begin{equation}\label{e18}
  d_j - m_j-\eta/2=0,
\end{equation}
where $d_j$ and $m_j$ are the $j-th$ element of $\mathbf{d}$ and $\mathbf{m}$ respectively. According to the constraint $\mathbf{1}_n\mathbf{d}=k$ in Problem (\ref{e15}), then 
\begin{equation}\label{e19}
  \eta = 2(k-\mathbf{1}_n^{\top}\mathbf{m})/n.
\end{equation}  
Hence, 
\begin{equation}\label{e20}
  \mathbf{d} = \mathbf{m} + \mathbf{1}_n(k-\mathbf{1}_n^{\top}\mathbf{m})/n.
\end{equation}
By summarizing the above computations, 
\begin{equation}\label{e21}
  \mathbf{D}^{h+1} = \mathbf{M} + \mathbf{Diag}\big(\mathbf{1}_n(k-\mathbf{1}_n^{\top}diag(\mathbf{M}))/n\big),
\end{equation}
where $\mathbf{Diag}\big(\mathbf{1}_n(k-\mathbf{1}_n^{\top}diag(\mathbf{M}))/n\big)$ is a diagonal matrix with its diagonal vector being $\mathbf{1}_n(k-\mathbf{1}_n^{\top}diag(\mathbf{M}))/n$.
\par \textbf{5. Fix other variables and update $\mathbf{E}$.} From Eq. (\ref{e10}), it could be easily obtained as follows:
\begin{equation}\label{e22}
  \begin{array}{l}
  \mathbf{E}_{h+1} =\arg\min_{\mathbf{E}} \lambda\|\mathbf{E}\|_{2,1} + <\mathbf{Y}_1^h,\mathbf{X}-\mathbf{XZ}^{h+1}-\mathbf{E}>\\
  +\mu^h\|\mathbf{X}-\mathbf{XZ}^{h+1}-\mathbf{E}\|_F^2\\
  =\arg\min_{\mathbf{E}} \lambda\|\mathbf{E}\|_{2,1} + \mu^h/2\|\mathbf{X}-\mathbf{XZ}^{h+1}-\mathbf{E}+\mathbf{Y}^h_1/\mu^h\|_F^2.
  \end{array}
\end{equation}

The above problem could be solved by following the Lemma presented in \cite{RN1710,conf_2}.
\par \textbf{Lemma 1.} Let $\mathbf{Q}=[\mathbf{q}_1,\mathbf{q}_2,\cdots,\mathbf{q}_i,\cdots]$ be a given matrix. If the optimal solution to
\begin{equation}\label{e22-1}
  \min_{\mathbf{P}}\alpha\|\mathbf{P}\|_{2,1} + \frac{1}{2}\|\mathbf{P}-\mathbf{Q}\|_F^2
\end{equation}
is $\mathbf{P}^*$, then the $i$-th column of $\mathbf{P}^*$ is
\begin{equation}\label{e22-2}
 \mathbf{P}^*(:,i)=\left\{
    \begin{array}{ll}
      \frac{\|\mathbf{q}_i\|_2-\alpha}{\|\mathbf{q}_i\|_2}\mathbf{q}_i, & \hbox{if $\alpha<\|\mathbf{q}_i\|_2$;} \\
      0, & \hbox{oterwise.}
    \end{array}
  \right.
\end{equation}

\par \textbf{6. Fix other variables and update parameters.} The precise updating schemes for the parameters existed in Eq. (\ref{e10}) are summarized as follows:
\begin{equation}\label{e23}
  \begin{array}{l}
    \mathbf{Y}_1^{h+1} = \mathbf{Y}^h_1+\mu^h(\mathbf{X}-\mathbf{XZ}^{h+1}-\mathbf{E}^{h+1}), \\
    \mathbf{Y}_2^{h+1} = \mathbf{Y}^h_2+\mu^h(\mathbf{S}^{h+1}-\mathbf{C}^{h+1}), \\
    \mathbf{Y}_3^{h+1} = \mathbf{Y}^h_3+\mu^h(\mathbf{1}_n^{\top}\mathbf{C}^{h+1}-\mathbf{1}_n^{\top}), \\
    \mathbf{Y}_4^{h+1} = \mathbf{Y}^h_4+\mu^h(\mathbf{S}^{h+1}-\mathbf{D}^{h+1}), \\
    \mu^{h+1} = \min(\mu_{max},\rho\mu^h),
  \end{array}
\end{equation}
where $\mu_{max}$ and $\rho$ are two given positive parameters.

\subsection{Algorithm}
We summarize the algorithmic procedure of IDR in \textbf{Algorithm 1}. For a data set, once the solutions to IDR are obtained, we use $\mathbf{Z}$ and $\mathbf{S}$ to define two affinity graphs $\mathbf{G}_1$ and $\mathbf{G}_2$ as $[\mathbf{G}_1]_{ij}=\big(|[\mathbf{Z}]_{ij}|+|[\mathbf{Z}^{\top}]_{ij}|\big)/2$ and $[\mathbf{G}_2]_{ij}=\big(|[\mathbf{S}]_{ij}|+|[\mathbf{S}^{\top}]_{ij}|\big)/2$. Then N-cut is consequently performed on the two graphs to get two segmentation results. Finally, the best one would be chosen as the final result.

\begin{algorithm}
  \small
  \renewcommand{\algorithmicrequire}{\textbf{Input:}}
  \renewcommand\algorithmicensure {\textbf{Output:}}
  \caption{Idempotent representation (IDR)}
  \begin{algorithmic}[1]
  \REQUIRE ~~\\
  Data set $\mathbf{X}=[\mathbf{x}_{1},\mathbf{x}_{2},\cdots,\mathbf{x}_{n}]\in \mathcal{R}^{d\times n}$ with each column has unit $l_2$ norm, parameters $\gamma,\lambda$, the number of subspaces $k$, the maximal number of iteration $Maxiter$;
  \ENSURE ~~\\
  The coefficient matrix $\mathbf{Z}^*$ and $\mathbf{S}^*$ and the noise term $\mathbf{E}^{*}$;
  \STATE Initialize the parameters, i.e., $h=0,\mu^h=10^{-6}, \mu_{max}=10^{4}, \rho=1.1, \varepsilon=10^{-7}$ and $\mathbf{Y}_1^h=\mathbf{Y}_2^h=\mathbf{Y}_3^h=\mathbf{0},\mathbf{Z}^h=\mathbf{S}^h=\mathbf{C}^h=\mathbf{D}^h=\mathbf{0}$.
  \WHILE {$\|\mathbf{S}^h-\mathbf{C}^h\|_{\infty}>\varepsilon$, $\|\mathbf{S}^h-\mathbf{D}^h\|_{\infty}>\varepsilon$, $\|\mathbf{1}_n^{\top}\mathbf{C}^h-\mathbf{1}^{\top}_n\|_{\infty}>\varepsilon$ and  $h<Maxiter$}
  \STATE $h = h + 1$;
  \STATE Update $\mathbf{Z}^{h+1} = \big(2\mathbf{I}_n + \lambda\mathbf{X}^{\top}\mathbf{X}\big)^{-1}\big(\mathbf{S}^h +\lambda\mathbf{X}^{\top}\mathbf{X}\big)$;
  \STATE Update $\mathbf{S}^{h+1}  = \big(2\mathbf{Z}^{h+1}+\mu^h\mathbf{C}^h-\mathbf{Y}^h_2+\mu^h\mathbf{D}^h-\mathbf{Y}^h_4\big)\big((2+2\mu^h)\mathbf{I}_n+2\gamma(\mathbf{I}_n-\mathbf{C}^h)(\mathbf{I}_n-\mathbf{C}^h)^{\top}\big)^{-1}$. Then let $\mathbf{S}^{h+1}=\max(\mathbf{0},\mathbf{S}^{h+1})$ and $\mathbf{S}^{h+1} =\big(\mathbf{S}^{h+1} + (\mathbf{S}^{h+1})^{\top}\big)/2$;
  \STATE Update $\mathbf{C}^{h+1} = \big(2\gamma(\mathbf{S}^{h+1})^{\top}\mathbf{S}^{h+1}+\mu^h(\mathbf{I}_n + \mathbf{1}_n\mathbf{1}_n^{\top})\big)^{-1}\big(2\gamma(\mathbf{S}^{h+1})^{\top}\mathbf{S}^{h+1}+\mathbf{Y}_2^h -\mathbf{1}_n\mathbf{Y}_3^h+\mu^h(\mathbf{S}^{h+1}+\mathbf{1}_n\mathbf{1}_n^{\top})\big)$. 
  \STATE Update $\mathbf{D}^{h+1} = \mathbf{M} + \mathbf{Diag}\big(\mathbf{1}_n(k-\mathbf{1}_n^{\top}diag(\mathbf{M})/n\big)$, where $\mathbf{M}=\mathbf{S}^{h+1}+\mathbf{Y}_3^{h}$;
  \STATE Update $\mathbf{E}^{h+1}$ by solving Problem (\ref{e22});
  \STATE Update $\mathbf{Y}_1^{h+1},\mathbf{Y}_2^{h+1},\mathbf{Y}_3^{h+1}$ and $\mu^{h+1}$ by using Eq. (\ref{e23}). 
  \ENDWHILE\label{code:recentEnd}
  \RETURN $\mathbf{Z}^{*}=\mathbf{Z}^{h}, \mathbf{S}^{*}=\mathbf{S}^{h}$.
  \end{algorithmic}
  \end{algorithm}

\section{Further analyses}
\label{sec4}
\subsection{Complexity analysis}
We can see that the complexity of \textbf{Algorithm 1} is mainly determined by the updating of five variables $\mathbf{Z,S,C,D,E}$. In each iteration, these variables all have closed form solutions. For updating $\mathbf{Z,S,C}$, it needs to compute the pseudo-inverse of an $n\times n$ matrix, hence the computation burden is $O(n^3)$. For updating $\mathbf{D}$, it takes $O(n^2)$ to compute the multiplier of an  $n\times n$ matrices. And for updating $\mathbf{E}$ by using \textbf{Lemma 1}, its time cost is $O(n)$. Hence, the time complexity of \textbf{Algorithm 1} in each iteration taken together is $O(n^3)$. In our experiments, the number of iterations of \textbf{Algorithm 1} are all less than $500$, hence its total complexity is $O(n^3)$.

\subsection{Convergence analysis}
Then we present a theoretical convergence proof of the proposed \textbf{Algorithm 1}. 
\par \emph{Proposition 1}: \textbf{Algorithm 1} is convergent and the sequence $\{\mathbf{Z}^h,\mathbf{S}^h,$ $\mathbf{C}^h,\mathbf{D}^h,\mathbf{E}^h\}$ generated by \textbf{Algorithm 1} would convergent to a stationary point.
\par \emph{Proof}: \textbf{Algorithm 1} aims to minimize the Lagrangian function of Eq. (\ref{e10}) by alternately updating the variables $\mathbf{Z,S,C,D,E}$. Firstly, from the updating rule of $\mathbf{Z}^{h+1}$ in Eq. (\ref{e11}), we have
  \begin{equation}\label{e24}
  \mathbf{Z}^{k+1}=\arg\min_{\mathbf{Z}} \mathfrak{L} (\mathbf{Z}^h,\mathbf{S}^h,\mathbf{C}^h,\mathbf{D}^h,\mathbf{E}^h).
  \end{equation} 
  Note that $\mathfrak{L} (\mathbf{Z}^h,\mathbf{S}^h,\mathbf{C}^h,\mathbf{D}^h,\mathbf{E}^h)$ is $\beta$-strongly convex w.r.t. $\mathbf{Z}$. The following inequality holds:
  \begin{equation}\label{e25}
    \begin{array}{l}
      \mathfrak{L} (\mathbf{Z}^{h+1},\mathbf{S}^h,\mathbf{C}^h,\mathbf{D}^h,\mathbf{E}^h)
    \leq \mathfrak{L} (\mathbf{Z}^{h},\mathbf{S}^h,\mathbf{C}^h,\mathbf{D}^h,\mathbf{E}^h)\\-\beta/2\|\mathbf{Z}^{h+1}-\mathbf{Z}^h\|_F^2.
    \end{array}
    \end{equation} 
Here we use Lemma B.5 in \cite{DBLP:conf/icml/Mairal13}.
\par Secondly, according to updating schemes for the rest variables, it could be found that these variables, namely $\mathbf{S,C,D,E}$, have the similar properties of $\mathbf{Z}$. Hence, the corresponding inequalities of the variables similar to (\ref{e25}) would hold. By adding these inequalities, we have
 \begin{equation}\label{e26}
   \begin{array}{l}
     \mathfrak{L} (\mathbf{Z}^{h+1},\mathbf{S}^{h+1},\mathbf{C}^{h+1},\mathbf{D}^{h+1},\mathbf{E}^{h+1})\leq  \mathfrak{L} (\mathbf{Z}^{h},\mathbf{S}^h,\mathbf{C}^h,\mathbf{D}^h,\mathbf{E}^h)\\-\beta/2\Big(\|\mathbf{Z}^{h+1}-\mathbf{Z}^h\|_F^2+\|\mathbf{S}^{h+1}-\mathbf{S}^h\|_F^2
      +\|\mathbf{C}^{h+1}-\mathbf{C}^h\|_F^2\\
      +\|\mathbf{D}^{h+1}-\mathbf{D}^h\|_F^2+\|\mathbf{E}^{h+1}-\mathbf{E}^h\|_F^2\Big).
   \end{array}
 \end{equation} 
Hence, $\mathcal{L}(\mathbf{Z}^{h},\mathbf{S}^{h},\mathbf{C}^{h},\mathbf{D}^{h},\mathbf{E}^{h})$ is monotonically decreasing and thus it is upper bounded. This implies that $\{\mathbf{Z}^h,\mathbf{S}^h,$ $\mathbf{C}^h,\mathbf{D}^h,\mathbf{E}^h$ are also bounded. Now, summing inequality (\ref{e26}) over $h=1,2,\cdots$, we have
  \begin{equation}\label{e27}
  \begin{array}{l}
    \sum_{k=1}^{+\infty}\frac{\beta}{2}\Big(\|\mathbf{Z}^{h+1}-\mathbf{Z}^h\|_F^2+\|\mathbf{S}^{h+1}-\mathbf{S}^h\|_F^2    \\+\|\mathbf{C}^{h+1}-\mathbf{C}^h\|_F^2
    +\|\mathbf{D}^{h+1}-\mathbf{D}^h\|_F^2+\|\mathbf{E}^{h+1}-\mathbf{E}^h\|_F^2\\
    \leq \mathcal{L}(\mathbf{Z}^{0},\mathbf{S}^{0},\mathbf{C}^{0},\mathbf{D}^{0},\mathbf{E}^{0}\Big).
  \end{array}
  \end{equation}
This implies when $h\rightarrow+\infty$,
  \begin{equation}\label{e28}
   \begin{array}{l}
    \mathbf{Z}^{h+1}-\mathbf{Z}^h\rightarrow 0,\\
    \mathbf{S}^{h+1}-\mathbf{S}^h\rightarrow 0,\\
    \mathbf{C}^{h+1}-\mathbf{C}^h\rightarrow 0,\\
    \mathbf{D}^{h+1}-\mathbf{D}^h\rightarrow 0,\\
    \mathbf{E}^{h+1}-\mathbf{E}^h\rightarrow 0.
   \end{array} 
  \end{equation}
Moreover, according to the definition, clearly $  \mathfrak{L} (\mathbf{Z}^{h},\mathbf{S}^{h},\mathbf{C}^{h},\mathbf{D}^{h},\mathbf{E}^{h})\geq 0$. Therefore, the convergence of \textbf{Algorithm 1} is guaranteed and the sequence $\{\mathbf{Z}^{h},\mathbf{S}^{h},\mathbf{C}^{h},\mathbf{D}^{h},\mathbf{E}^{h}\}$ would convergent to a stationary point of Eq. (\ref{e9}). 

\subsection{Comparative analysis with related algorithms}
We now discuss the relationships between IDR and some related algorithms.
\subsubsection{Comparative analysis with membership representation (MR)}
As we mentioned in Section \ref{sec1}, MR also proposes to learn a normalized membership matrix as a reconstruction coefficient matrix \cite{DBLP:conf/cvpr/LeeLLK15}. However, MR is a cascade model which consists of three steps:
\par Firstly, an initial coefficient matrix $\mathbf{W}$ is leaned by using SSC or LRR.
\par Secondly, a membership matrix $\mathbf{M}$ is constructed by solving the following problem:
\begin{equation}\label{e30}
  \begin{array}{ll}
    \min_{\mathbf{M}}&\|\mathbf{W}-\mathbf{W}\bigodot\mathbf{M}\|_1+\lambda\|\mathbf{M}\|_F^2\\
    s.t. & diag(\mathbf{M}) = \mathbf{1}_n,\mathbf{M}\geq 0,\mathbf{M}\succeq \mathbf{0},
  \end{array}
\end{equation}
where $\mathbf{M}\succeq \mathbf{0}$ requires $\mathbf{M}$ to be a positive semi-definite, $\lambda>0$ is a positive parameter. 
\par Thirdly, after $\mathbf{M}$ is obtained, a normalized membership matrix $\mathbf{Z}$ is achieved by optimizing the following problem:
\begin{equation}\label{e31}
  \begin{array}{ll}
    \min_{\mathbf{Z}}&Tr(\mathbf{Z})\\
    s.t. & \mathbf{1}_n^{\top}\mathbf{Z}=\mathbf{1}_n^{\top},\mathbf{Z}\geq 0,\mathbf{Z}\succeq \mathbf{0},\\
    &<\mathbf{H},\mathbf{Z}>\leq c,
  \end{array}
\end{equation}
where $\mathbf{H} = \mathbf{1}_n\mathbf{1}_n^{\top}-\mathbf{M}$ and $c = \beta\|\mathbf{H}\|_1/n$. $\beta>0$ is a manually set constant. We could see that the symmetric constraint of $\mathbf{Z}$ is omitted in the above problem. Hence, the coefficient matrix found by MR may not be close to a normalized membership matrix. 
\par Besides the computation for finding the initial coefficient matrix, Problem (\ref{e30}) and (\ref{e31}) are also needed to solve by using ALM method. Clearly, MR is very time-consuming. 
\par Additionally, it can be seen that the performances of MR depends on the learned initial coefficient matrices. The value of the parameter in SSC or LRR will influence its performances. How to choose an initial coefficient matrix is not reported discussed in \cite{DBLP:conf/cvpr/LeeLLK15}. Moreover, the three hyper-parameters existed in MR will make the tuning of the parameters be difficult. 
\subsubsection{Comparative analysis with doubly stochastic subspace clustering (DSSC) \cite{DBLP:journals/corr/abs-2011-14859}}
Based on the descriptions in Section \ref{sec3.1}, it could be seen that the normalized membership matrix obtained by IDR is a special case of doubly stochastic matrix. Recently, Lim et al. devised a doubly stochastic subspace clustering (DSSC) algorithm \cite{DBLP:journals/corr/abs-2011-14859} which pursuits a doubly stochastic coefficient matrix. The objective of DSSC could be expressed as follows:
\begin{equation}\label{e32}
  \begin{array}{ll}
    \min_{\mathbf{Z,A}} & \gamma\|\mathbf{Z}\|_1 +\lambda/2\|\mathbf{Z} -\eta \mathbf{A}\|_F^2+1/2\|\mathbf{X-XZ}\|_F^2\\
    s.t. & diag(\mathbf{Z})=0, \mathbf{A} \in \Psi_n,
  \end{array}
\end{equation}
where $\gamma,\lambda,\eta>0$ are three parameters and $\mathbf{A} \in \Psi_n$ means $\mathbf{A}$ is an $n\times n$ doubly stochastic matrix. Namely, $\mathbf{A}$ satisfies the conditions presented in Eq. (\ref{e6}). By using two different strategies to solve the above problem, two different models, joint DSSC (J-DSSC) and approximation DSSC (A-DSSC), are presented. Among them, A-DSSC is a two-step algorithm which firstly uses LSR or EN to get an initial coefficient matrix and computes a doubly stochastic matrix consequently. On the other hand, the computation burden of J-DSSC is high, because in each iteration of J-DSSC, two intermediate $n\times n$ matrices should be iteratively updated by using linear alternating direction method of multipliers (ADMM) \cite{DBLP:journals/jscic/Ma16}. Moreover, we also could see that DSSC has three hyper-parameters which will also leads the difficulties in parameters adjustment. 
\subsubsection{Comparative analysis with self-representation constrained LRR (SRLRR) \cite{RN2445}}
The idempotent constraint of coefficient matrix is firstly proposed in our previously work in \cite{RN2445}. The SRLRR defines the following problem:
\begin{equation}\label{en33}
  \begin{array}{ll}
    \min_{\mathbf{Z,E}} & \|\mathbf{Z}\|_* +\gamma\|\mathbf{Z} -\mathbf{Z}^2\|_F^2+\lambda\|\mathbf{E}\|_{2,1}\\
    s.t. & \mathbf{X}=\mathbf{XZ}+\mathbf{E}, \mathbf{1}_n^{\top}\mathbf{Z} =\mathbf{1}_n^{\top}.
  \end{array}
\end{equation}
The main problem existed in SRLRR is that we have not build solid theoretical connections between $\mathbf{Z}$ and a normalized membership matrix. The nuclear norm minimization and the affine constraint (i.e., $\mathbf{1}_n^{\top}\mathbf{Z} =\mathbf{1}_n^{\top}$) \cite{RN2572} in SRLRR are used to avoid $\mathbf{Z}$ to degenerate to $\mathbf{I}_n$ or $\mathbf{0}_{n\times n}$. This is totally different from IDR.
\par Based on these comparisons, we could see the existing subspace clustering methods, which also aim to seek normalized membership matrices or doubly stochastic matrices, will all use certain existing regularizers of $\mathbf{Z}$. IDR presents a much different method for tacking subspace clustering problems.

\section{Experiments}
\label{sec5}
\subsection{Experiment setup}
\subsubsection{Datasets}
Both synthetic and real world data sets are applied in our experiments to verify the effectiveness of IDR. Four benchmark databases including Hopkins 155 motion segmentation data set \cite{DBLP:conf/cvpr/TronV07},  ORL face image database \cite{DBLP:conf/wacv/SamariaH94}, AR face image database \cite{ARface} and MNIST handwritten digital database\footnote{http://yann.lecun.com/exdb/mnist/} are used for evaluation.
\subsubsection{Comparison methods} The representative and close related algorithms such as SSC \cite{conf_1},  LRR \cite{conf_2,RN1710}, LSR \cite{Lu:2012}, BDR \cite{RN2485}, MR \cite{DBLP:conf/cvpr/LeeLLK15} and DSSC \cite{DBLP:journals/corr/abs-2011-14859} would be used for comparison\footnote{We provide the Matlab codes of IDR, MR and DSSC on \url{https://github.com/weilyshmtu/Learning-idempotent-representation-for-subspace-segmentation}. And the Matlab codes for SSC and LRR could be found on \url{http://www.vision.jhu.edu/code/} and \url{http://sites.google.com/site/guangcanliu/} respectively. The Matlab codes of LSR and BDR could be found on https://canyilu.github.io/code/.}. All the experiments are conducted on a Windows-based machine with an Intel i7-4790 CPU with 20-GB memory and MATLAB R2017b. 
\subsubsection{Parameters Setting}
\label{sec5.1.3}
Because the value of parameters will influence the performance of the evaluated algorithms, for each compared method, we will tune all the parameters by following the suggestions in corresponding references and retain those with the best performance on each data set. The chosen parameter settings for all algorithms are given in Table \ref{t2}. Especially, for MR, when SSC or LRR is used to achieve the initial coefficient matrix, the parameter corresponding to the two algorithms would be chosen in $[0.001,0.01,0.1,1,5,10]$ according to the description in \cite{DBLP:conf/cvpr/LeeLLK15}. 
\begin{table*}
  \begin{center}
  
  \caption{Parameters searched over for different methods.}\label{t2}
  \scriptsize
  \begin{tabular}{c|l}
    \hline
    Methods & Parameters \\\hline
    SSC     & $\lambda \in \{0.0001,0.001,0.01,0.1,1,10,20,50,100,200,500,600,800,1000\}$\\\hline
    LRR     & $\lambda \in \{0.0001,0.001,0.01,0.05,0.1,0.2,0.5,1,2,5,8,10,15,20,50\}$\\\hline
    LSR     & $\lambda \in \{0.0001,0.001,0.01,0.05,0.1,0.2,0.5,1,2,5,8,10,15,20,50\}$\\\hline
    BDR     & $\lambda \in \{0.001,0.01,0.05,0.1,0.2,0.5,1,2,3,5,8,10,15,20,50\}$,\\
            & $\gamma \in \{0.1,1,10,20,30,40,50,60,70,80\}$\\\hline
    MR      & $\lambda \in \{0.001,0.01,0.1,1,5,10\},\beta \in \{0.01,0.1,1,5,10\}$\\\hline
    DSSC(JDSSC)    & $\lambda \in \{0.01,0.25,1,25\}, \eta\in\{0.01,0.05,0.1,0.2\},\gamma\in\{0,0.01\}$\\ \hline
    DSSC(ADSSC)    & $\lambda \in \{0.1,1,10,25,50\}, \eta\in\{0.0005,0.001,0.01,0.025,0.05,0.1\},\gamma=0$\\\hline
    IDR & $\lambda,\gamma\in\{0.001,0.005,0.01,0.02,0.05,0.1,0.2,0.5,1,5,10,50,100,200\}$\\\hline
  \end{tabular}
\end{center}
\end{table*}
\subsubsection{Evaluation metrics} For all the evaluated algorithms, we use the obtained coefficient matrices to construct the affinity graphs without any post-processing. For the performance evaluation, we use segmentation accuracy (SA) or segmentation error (SE), which is defined as follows:
\begin{equation}\label{e33}
  \begin{array}{l}
    SA =\frac{1}{n}\sum_{i=1}^n\delta(s_i,f(t_i)),\\
    SE = 1 - SA,
  \end{array}
\end{equation}
where $s_i$ and $t_i$ represent the ground truth and the output label of the $i$-th point respectively, $\delta(x, y) = 1$ if $x = y$, and $\delta(x, y) = 0$ otherwise, and $f(t_i)$ is the best mapping function that permutes clustering labels to match the ground truth labels.
\subsection{Experiments on a synthetic data set}
We generate $5$ subspaces each of dimension $d = 5$ in an ambient space of dimension $D = 20$. We sample $50$ data points from each subspace and construct a $D\times 250$ data matrix $\mathbf{X}$. Moreover, a certain percentage $p = 0-100\%$ of the data vectors are chosen randomly to add Gaussian noise with zero mean and variance $0.3\|\mathbf{x}\|_2^2$. Finally, the evaluated algorithms are used to segment the data into $5$ subspaces. For a certain $p$, the experiments would repeat $20$ trials. Therefore, there would be total $220$ subspace clustering tasks. 
\par
\begin{figure}
  \centering
  \includegraphics[width=0.5\textwidth]{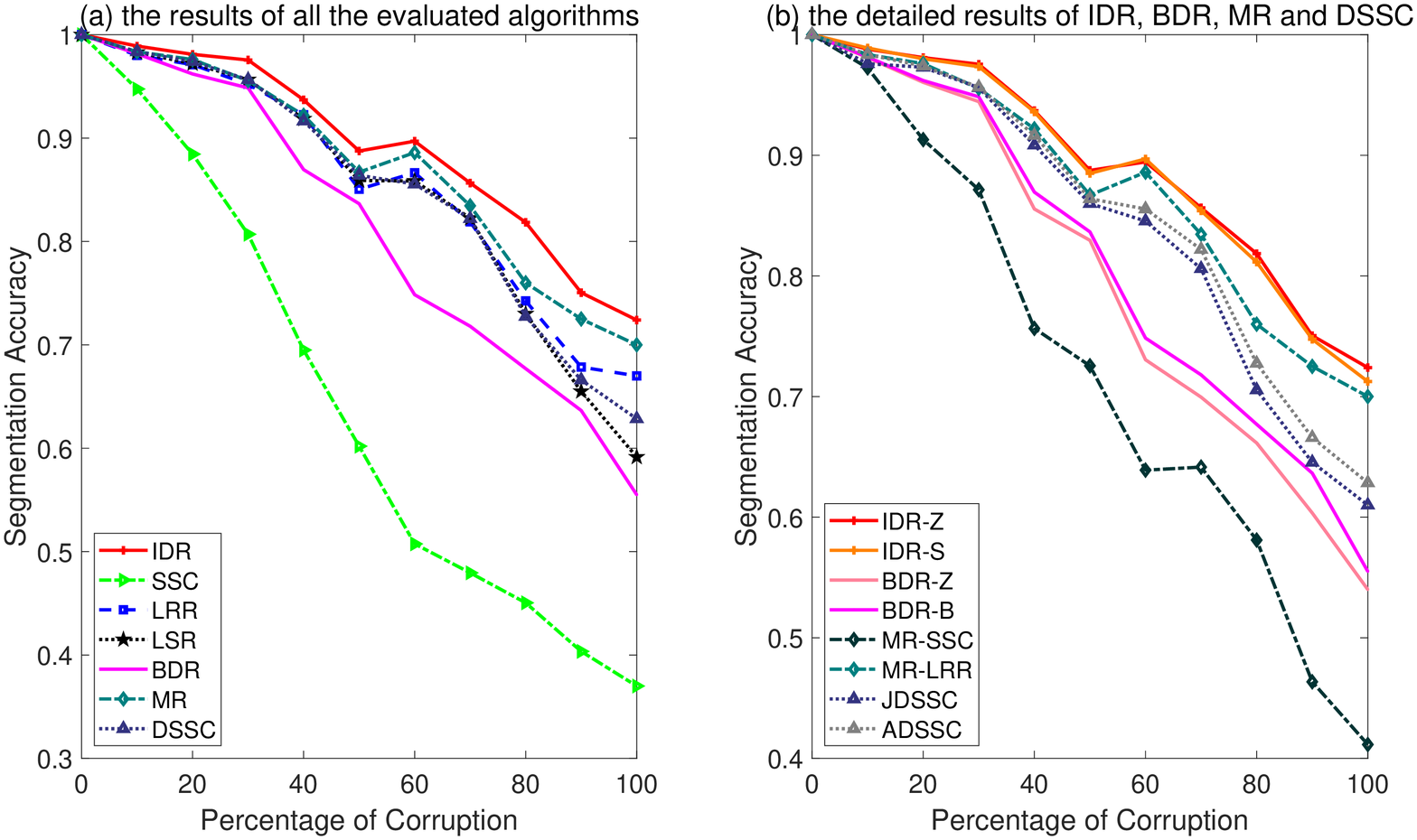}\\
  \caption{The segmentation accuracies of each method versus variation the range of corruption. (a) The best results of the evaluated algorithms, (b) the detailed results of IDR, BDR, MR and DSSC, where BDR-Z denotes the results obtained by the reconstruction coefficient matrix $\mathbf{Z}$ and BDR-B indicates the results obtained by the intermediate matrix introduced in BDR problem.}\label{f2}
\end{figure}

 Actually, the similar experiments could be found in some existing references \cite{conf_2,RN2328}. But in these experiments the parameters of corresponding evaluated algorithm would be fixed when the algorithm is performed on each sub-database. Then by changing the parameter(s), the best results with certain parameter(s) would be finally selected. However, performing subspace clustering on sub-database should be viewed as a sole segmentation task. In our experiments, we hence will let the parameter(s) of each algorithm vary in the corresponding interval sets in Table \ref{t2} and record the highest segmentation accuracies of the evaluated algorithms on each sub-database. Then the mean of these highest segmentation accuracies (averaged from 20 random trials) of each algorithm versus variation of the percentage of corruption are reported in Fig. \ref{f2}. 
\par In addition, IDR and BDR can produce two coefficient matrices to compute the clustering results. By using different methods (SSC and LRR) to construct initial coefficient matrices, MR could obtain two different results. Based on different strategies, DSSC has two sub-models, namely JDSSC and ADSSC. Hence, we plot the accuracies of all the algorithms by selecting the better ones of the corresponding two results in Fig \ref{f2}(a). The detailed segmentation accuracies of IDR and BDR by using two different coefficient matrices, and the results of MR based on SSC and LRR (denoted as MR-SSC and MR-LRR) as well as JDSSC and ADSSC are ploted in Fig. \ref{f2}(b). 
\par From Fig. \ref{f2}(a), we can see that 1) IDR constantly achieves the best results; 2) the performances of LRR, LSR, MR and DSSC are closed to each other, when the percentage of corruption is smaller than $50\%$ ; 3) when the percentage of corruption is larger than $50\%$, MR dominates LRR, LSR and DSSC; 4) SSC is inferior to other algorithms. 
\par From Fig. \ref{f2}(b), it can be seen that 1) the results obtained by two different coefficient matrices corresponding to IDR and BDR respectively are closed to each other; 2) the performances of JDSSC and ADSSC are also similar to each other; 3) However, the results of MR-LRR are much better than those of MR-SSC. This means that the performance of MR relies on the initial coefficient matrices.

%

\par
\begin{figure*}
  \centering
  \includegraphics[width=0.7\textwidth]{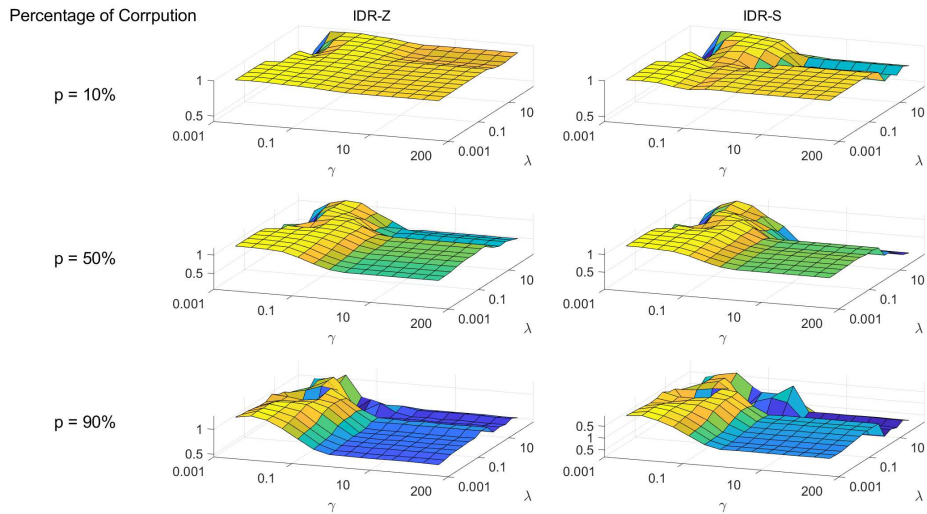}\\
  \caption{The segmentation accuracy against the variation of parameters of IDR. The vertical axis in each figure denotes the subspace clustering accuracy. The three sub-figures in the left column show the segmentation accuracies obtained by $\mathbf{Z}$ changed with parameters, and the three sub-figures in the right column record the segmentation accuracies obtained by $\mathbf{S}$ changed with parameters. The first, second and third rows present the segmentation accuracies of IDR on the sub-databases with corruption percentage equals $10\%,50\%,90\%$ respectively.}\label{f3}
\end{figure*}

 In order to show the sensitivity of IDR to its two parameters $\gamma$ and $\lambda$, we report the segmentation accuracies of IDR changed with the values of parameters. The sub-databases with $p=10\%,50\%,90\%$ are now used. Then the mean of segmentation accuracies against the pairs of parameters are illustrated in Fig. \ref{f3}.  
\par From Fig. \ref{f3}, we could see that 1) the performance of IDR is stable when the parameters varied in relative large intervals; 2) when the corruption percentage is low, IDR is insensitive to $\gamma$. 
However, when the corruption percentage is high, small $\gamma$ and $\lambda$ could help IDR to achieve good results. We believe that when a data set is clean, a normalized membership reconstruction coefficient matrix is easily to get, so the idempotent constraint could also be satisfied. However, when most data samples in the data set are corrputed, the normalized membership reconstruction coefficient matrix is difficult to obtain. Hence, in such situations, the corresponding parameter $\gamma$ should be small.

\subsection{Experiments on Hopkins 155 data set}
\par
\begin{figure*}
  \centering
  \includegraphics[width=0.7\textwidth]{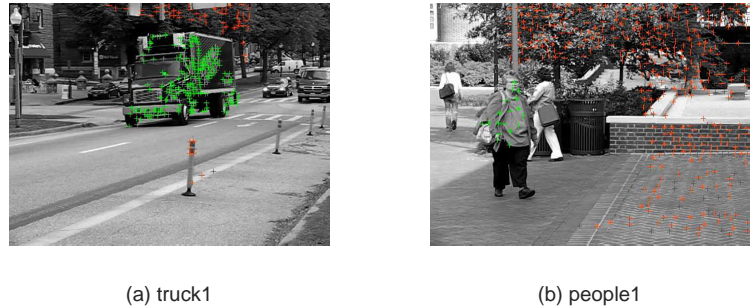}\\
  \caption{The sample images from Hopkins 155 database.}\label{f4}
\end{figure*}

 Hopkins155 database is a well-known benchmark database to test the performances of subspace clustering algorithms. It consists of 120 sequences of two motions and 35 sequences of three motions. Each sequence is a sole clustering task and there are 155 clustering tasks in total. The features of each sequence are extracted and tracked along with the motion in all frames, and errors are manually removed for each sequence. We illustrate the sample images from Hopkins 155 database in Fig.~\ref{f4}.
\par We performed the experiments with the original data matrix and projected data matrix in $4s-$dimensional subspace\footnote{$s$ is the number of subspaces.} obtained by using principal component analysis (PCA) \cite{DBLP:books/lib/DudaHS01}. Then the segmentation errors (i.e., $SE= 1-SA$) of each evaluated algorithm are computed on each sequence.
\par Firstly, we collect all the best results of each algorithm obtained on 155 sequences with the parameters changing in the given intervals. And the mean, median and std. (standard variance) of the results are reported in Table \ref{t3} and \ref{t4}. From the two tables, we could see that 1) IDR achieves best results in these experiments; 2) BDR and LSR also achieve competitive results; 3) MR-LRR and MR-SSC do not conquer their corresponding classical methods SSC and LRR. This means the post-processing on the coefficient matrices in MR may not always enhance the performance of LRR and SSC; 5) JDSSC fails to achieve satisfying results.
\begin{table*}
  \scriptsize
  \begin{center}
  \caption{The segmentation errors ($\%$) and average computation time (sec.) of different algorithms on the Hopkins 155 database with the original data points. The best results of these algorithms are emphasized in bold.}\label{t3}
  \begin{tabular}{c|c|ccc|ccc|ccc}
    \hline
    \multirow{2}{*}{Methods} & Average time &\multicolumn{3}{c|}{2 motions}  &\multicolumn{3}{c|}{3 motions}  &\multicolumn{3}{c}{All motions} \\ 
 
                             &  (sec.)    &  mean  &      median  &         std.  &        mean  &      median  &         std.  &        mean  &      median  &            std. \\ \hline
                      IDR-Z  &  \multirow{2}{*}{$9.15$}  &  $\mathbf{0.25}$  &         $\mathbf{0}$  &      $\mathbf{1.15}$  &      $\mathbf{1.14}$  &       $0.20$  &      $\mathbf{2.11}$  &      $\mathbf{0.45}$  &         $\mathbf{0}$  &         $\mathbf{1.47}$ \\ 
                      IDR-S  &      & $0.50$  &        $\mathbf{0}$  &      $1.89$  &      $2.23$  &      $0.56$  &      $3.49$  &      $0.89$  &         $\mathbf{0}$  &         $2.44$ \\ \hline
                        SSC  &  $2.98$  &  $1.66$  &         $\mathbf{0}$  &      $5.13$  &      $5.29$  &      $1.46$  &      $7.35$  &      $2.48$  &         $\mathbf{0}$  &         $5.88$ \\ \hline
                        LRR  &  $9.378$ &    $1.15$  &         $\mathbf{0}$  &      $3.19$  &      $4.17$  &       $1.20$  &      $5.99$  &      $1.83$  &         $\mathbf{0}$  &         $4.17$ \\  \hline
                        LSR  & $\mathbf{0.03}$ &    $0.56$  &         $\mathbf{0}$  &      $2.18$  &      $1.94$  &      $0.21$  &      $4.12$  &      $0.87$  &         $\mathbf{0}$  &         $2.79$ \\ \hline
                      BDR-B  & \multirow{2}{*}{$5.50$}  &   $0.58$  &         $\mathbf{0}$  &      $2.78$  &      $2.72$  &         $\mathbf{0}$  &      $4.73$  &      $1.06$  &         $\mathbf{0}$  &         $3.42$ \\ 
                      BDR-Z  &     &  $0.6$  &         $\mathbf{0}$  &      $2.76$  &      $2.77$  &         $\mathbf{0}$  &       $5.10$  &      $1.09$  &         $\mathbf{0}$  &         $3.53$ \\\hline 
                   MR-SSC  &  $41.15$ &  $2.71$  &         $\mathbf{0}$  &      $6.56$  &      $9.22$  &      $6.05$  &      $9.18$  &      $4.33$  &      $0.21$  &         $7.78$  \\
                   MR-LRR  &  $43.29$ &    $1.39$  &         $\mathbf{0}$  &      $3.95$  &      $6.52$  &      $2.85$  &      $6.82$  &      $2.66$  &         $\mathbf{0}$  &         $5.28$ \\ \hline
                      JDSSC  & $16.29$ &  $12.51$  &     $11.45$  &     $10.54$  &     $24.09$  &     $25.06$  &     $11.62$  &     $15.13$  &     $14.48$  &         $11.8$ \\ 
                      ADSSC  &  $0.07$ &   $2.42$  &         $\mathbf{0}$  &      $5.67$  &      $8.74$  &      $5.37$  &      $9.65$  &      $3.85$  &         $\mathbf{0}$  &         $7.24$ \\ 
\hline
  \end{tabular}
  \end{center}
\end{table*}
    
\begin{table*}
  \scriptsize
  \begin{center}
  \caption{The segmentation errors ($\%$) and average computation time (sec.) of different algorithms on the  Hopkins 155 database with the $4s-$dimensional data points by applying PCA. The best results of these algorithms are emphasized in bold.}\label{t4}
  \begin{tabular}{c|c|ccc|ccc|ccc}
    \hline
    \multirow{2}{*}{Methods} & Average time &\multicolumn{3}{c|}{2 motions}  &\multicolumn{3}{c|}{3 motions}  &\multicolumn{3}{c}{All motions} \\ 
 
    &  (sec.)    &  mean  &      median  &         std.  &        mean  &      median  &         std.  &        mean  &      median  &            std. \\ \hline
   IDR-Z  & \multirow{2}{*}{$9.43$}   &   $\mathbf{0.30}$  &         $\mathbf{0}$  &      $\mathbf{1.24}$  &       $\mathbf{1.20}$  &         $\mathbf{0}$  &       $\mathbf{2.30}$  &       $\mathbf{0.50}$  &         $\mathbf{0}$  &         $1.58$ \\ 
   IDR-S  &     & $0.49$  &        $\mathbf{0}$  &      $1.76$  &      $2.16$  &      $0.56$  &      $3.27$  &      $0.86$  &         $\mathbf{0}$  &         $2.29$ \\  \hline
     SSC  &  $2.29$ &   $1.66$  &         $\mathbf{0}$  &      $5.13$  &      $5.29$  &      $1.46$  &      $7.35$  &      $2.48$  &         $\mathbf{0}$  &         $5.88$ \\  \hline
     LRR  &  $10.27$  &  $1.15$  &         $\mathbf{0}$  &      $3.19$  &      $4.17$  &       $1.20$  &      $5.99$  &      $1.83$  &         $\mathbf{0}$  &         $4.17$ \\  \hline
     LSR  &  $\mathbf{0.03}$  &  $0.56$  &         $\mathbf{0}$  &      $2.18$  &      $1.94$  &      $0.21$  &      $4.12$  &      $0.87$  &         $\mathbf{0}$  &         $2.79$ \\  \hline
   BDR-Z  &  \multirow{2}{*}{$4.41$}  &  $0.65$  &         $\mathbf{0}$  &      $2.82$  &      $2.91$  &         $\mathbf{0}$  &      $5.12$  &      $1.16$  &         $\mathbf{0}$  &         $3.59$ \\ 
   BDR-B  &    &  $0.65$  &         $\mathbf{0}$  &       $2.90$  &      $2.68$  &         $\mathbf{0}$  &      $5.03$  &      $1.11$  &         $\mathbf{0}$  &         $3.58$ \\  \hline
MR-SSC  & $39.89$   &  $4.62$  &         $\mathbf{0}$  &      $8.22$  &     $10.44$  &     $10.47$  &      $8.59$  &      $6.06$  &      $0.47$  &         $8.65$ \\  
MR-LRR  & $44.47$ &    $1.39$  &         $\mathbf{0}$  &      $3.93$  &      $6.72$  &      $2.85$  &      $6.93$  &      $2.71$  &         $\mathbf{0}$  &         $5.34$ \\ \hline
   JDSSC  & $14.18$   &  $12.51$  &     $11.45$  &     $10.54$  &     $24.09$  &     $25.06$  &     $11.62$  &     $15.13$  &     $14.48$  &         $11.8$ \\ 
   ADSSC  & $0.07$ &    $2.42$  &         $\mathbf{0}$  &      $5.67$  &      $8.76$  &      $5.37$  &      $9.65$  &      $3.85$  &         $\mathbf{0}$ &         $7.24$ \\ 
\hline
  \end{tabular}
  \end{center}
\end{table*}
\par Moreover, we also report the average computation time of each algorithm on the 155 motion sequences in Table \ref{t3} and \ref{t4}. Clearly, LSR and ADSSC are much efficient than other algorithms. The average computation time of IDR is close to that of LRR. Hence the computation burden of IDR is acceptable. We could see that MR is time-consuming.
\par
\begin{figure*}
  \centering
  \includegraphics[width=0.7\textwidth]{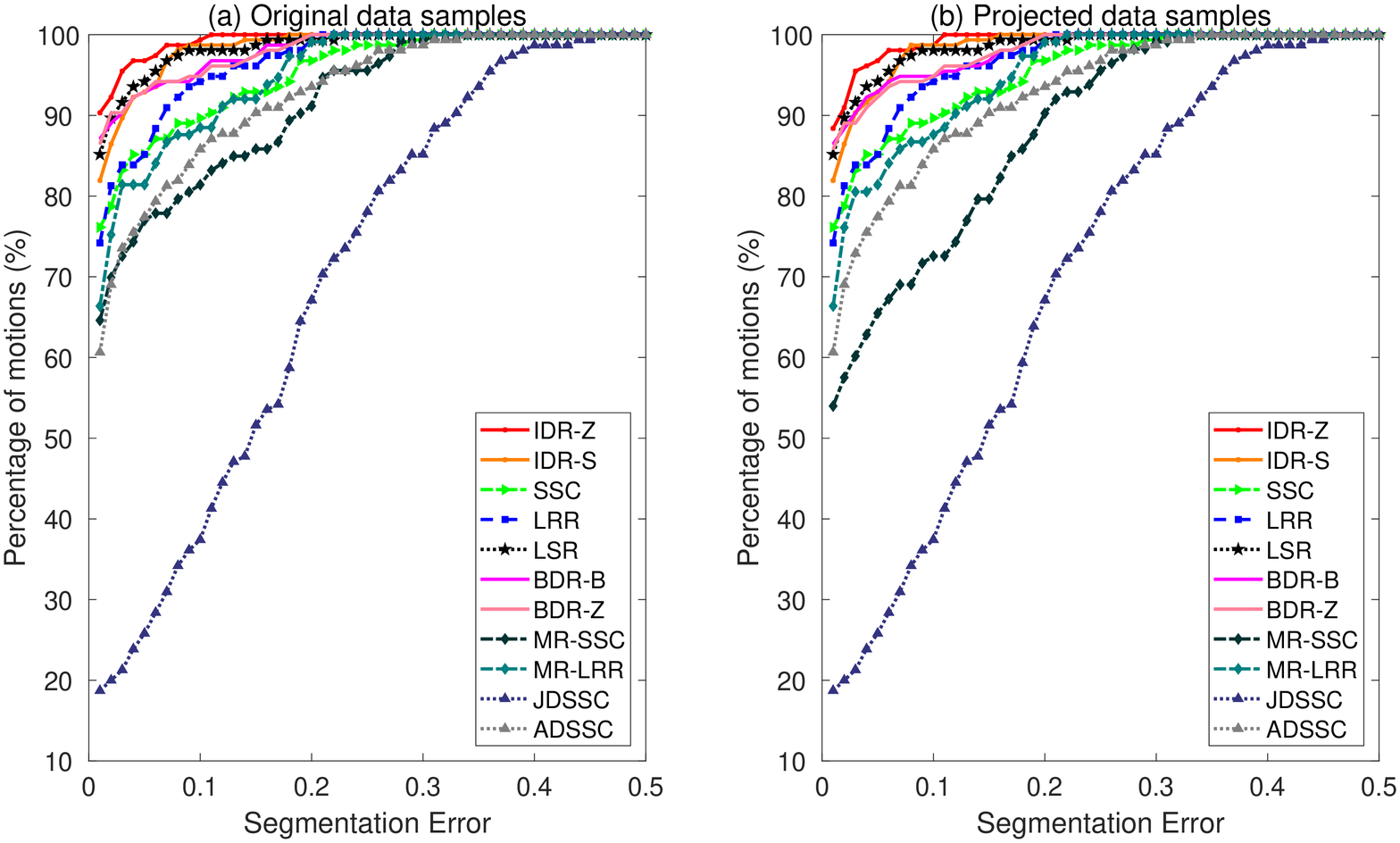}\\
  \caption{The percentage of motions versus the variation of segmentation error for each method. (a) The results obtained on the original data samples, (b) The results obtained on the projected data samples.}\label{f5}
\end{figure*}

 Secondly, we analyse the experimental results of each algorithm in another way. For each algorithm, we present the percentage of motions with the obtained SEs are less than or equal to a given percentage of segmentation error in Fig. \ref{f5}. We can see that the segmentation errors on all motions obtained by IDR-Z are all less than $0.2$. 

\par
\begin{figure*}
  \centering
  \includegraphics[width=0.7\textwidth]{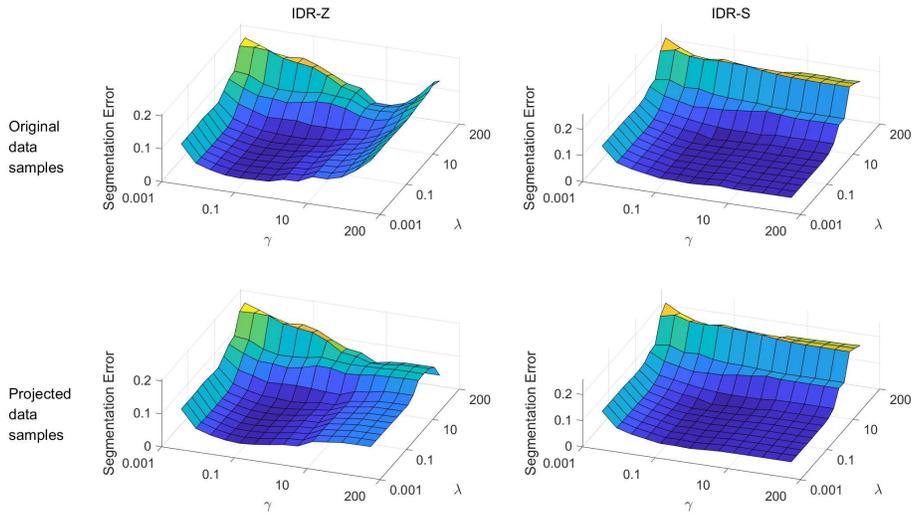}\\
  \caption{The segmentation accuracy against the variation of parameters of IDR on Hopkins 155 data set. Now the vertical axis in each figure denotes the subspace clustering error which is different from the experiments conducted on synthetic data sets. The two sub-figures in the left column show the segmentation errors obtained by $\mathbf{Z}$ changed with parameters, and the two sub-figures in the right column record the segmentation errors obtained by $\mathbf{S}$ changed with parameters. The first and second rows present the segmentation error of IDR on the original data sets and projected data sets respectively.}\label{f6}
\end{figure*}

 Finally, we also test the sensitivity of IDR to the parameters on Hopkins 155 database. For a fixed pair of $(\gamma,\lambda)$, we compute the segmentation error for all 155 segmentation tasks, then the mean of 155 segmentation accuracies could be achieved. Then by changing the values of $\gamma$ and $\lambda$, we illustrate the performance of IDR against to its parameters in Fig \ref{f6}.
Based on Fig. \ref{f6}, we still could see that IDR is insensitive to its parameters and small $(\gamma, \lambda)$ could help IDR to achieve better results.

\subsection{Experiments on face image databases}
We now perform the experiments on two benchmark face image databases, i.e., ORL database \cite{DBLP:conf/wacv/SamariaH94} and AR database \cite{ARface}. The brief information of the two databases is introduced as follows:
\par ORL database contains $400$ face images (without noise) of $40$ persons. Each individual has 10 different images. These images were taken at different times, varying the lighting, facial expressions (open/closed eyes, smiling/not smiling) and facial details (glasses/no glasses). In our experiments, each image is resized to $32\times 32$ pixels.
\par AR database consists of over $4000$ face images of $126$ individuals. For each individual, 26 pictures were taken in two sessions (separated by two weeks) and each section contains 13 images. These images include front view of faces with different expressions, illuminations and occlusions. In our experiments, each image is resized to $50\times 40$ pixels.
\par
\begin{figure}
  \centering
  \includegraphics[width=0.5\textwidth]{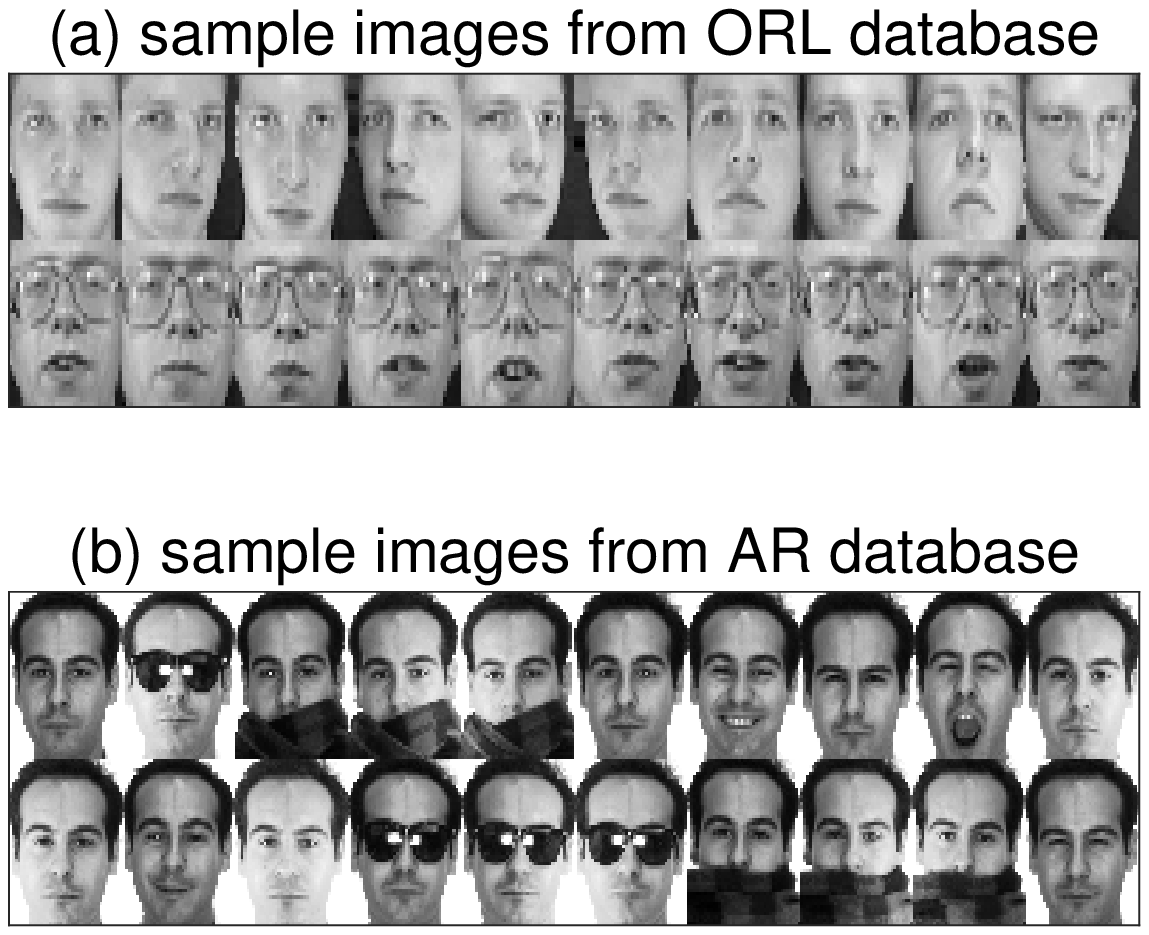}\\
  \caption{Sample images from ORL database and AR database.}\label{f7}
\end{figure}

 Moreover, the pixel value in each image belongs to the two databases lies in $[0,255]$. For efficient computation, we let each pixel value be divided by $255$, so that the pixel value of each image fell into $[0,1]$. This will not change the distribution of the original data sets. Some sample images from ORL and AR database are shown in Fig. \ref{f7}. 
\par We firstly randomly choose images of $q$ persons from the two databases. In ORL database, $q\in \{6,12,18,24,36\}$, and in AR database, $q\in \{4,8,12,16,20\}$. Then the performances of the evaluated methods are tested in these sub-databases. With the parameters varying, the highest clustering accuracy of each algorithm obtained on each sub-database is collected. These experiments are run $10$ trials, and the mean and standard variance of SAs obtained by each algorithm are reported in Table \ref{t5} and Table \ref{t6} respectively.

\begin{table*}
  \scriptsize
  \begin{center}
  \caption{Segmentation accuracies ($\%$) $\pm$ standard variances of evaluated algorithms on the sub-databases of ORL database. The best results (mean) of the algorithms are emphasized in bold.}\label{t5}
  \begin{tabular}{c|cccccc}
  \hline
     \multirow{2}{*}{Methods}  &                     \multicolumn{6}{c}{Number of persons}          \\\cline{2-7}
                               &                        $6$  &                       $12$  &                       $18$  &                       $24$  &                       $30$  &                          $36$       \\\hline
                               IDR-Z  &             $92.33\pm7.42$  &             $87.08\pm3.91$  &             $87.28\pm4.21$  &             $83.54\pm2.56$  &             $82.40\pm1.74$  &                $\mathbf{81.89\pm0.87}$        \\
                               IDR-S  &             $\mathbf{94.17\pm6.35}$  &             $\mathbf{89.50\pm1.63}$  &             $87.39\pm2.66$  &             $83.67\pm2.28$  &             $\mathbf{83.03\pm1.37}$  &                $81.81\pm0.85$        \\\hline
                          SSC  &             $89.83\pm7.22$  &             $80.25\pm5.81$  &             $80.44\pm5.12$  &             $79.00\pm2.57$  &             $77.93\pm2.88$  &                $77.36\pm1.89$        \\\hline
                          LRR  &             $89.00\pm7.04$  &             $82.50\pm5.61$  &             $84.67\pm4.08$  &             $80.71\pm2.88$  &             $79.43\pm1.85$  &                $77.75\pm1.26$        \\\hline
                          LSR  &             $91.67\pm5.88$  &             $86.08\pm3.95$  &             $86.33\pm3.47$  &             $82.92\pm3.11$  &             $81.60\pm1.14$  &                $80.69\pm1.52$        \\\hline
                        BDR-Z  &             $90.50\pm7.20$  &             $85.67\pm5.03$  &             $86.89\pm3.49$  &             $83.38\pm2.56$  &             $80.53\pm1.31$  &                $80.36\pm1.09$        \\
                        BDR-B  &             $91.83\pm7.13$  &             $86.08\pm2.97$  &             $\mathbf{88.28\pm3.22}$  &             $84.17\pm2.76$  &             $81.20\pm2.66$  &                $80.19\pm0.67$        \\\hline
                       MR-SSC  &             $91.67\pm6.67$  &             $87.17\pm6.59$  &             $86.00\pm3.48$  &             $\mathbf{84.46\pm2.22}$  &             $82.90\pm1.67$  &                $80.75\pm1.06$        \\
                       MR-LRR  &             $85.00\pm6.57$  &             $86.08\pm3.51$  &             $84.94\pm3.98$  &             $81.71\pm3.54$  &             $78.23\pm1.71$  &                $78.92\pm0.55$        \\\hline
                        JDSSC  &             $88.00\pm7.93$  &             $87.83\pm5.14$  &             $86.83\pm3.46$  &             $81.75\pm2.23$  &             $77.83\pm1.89$  &                $77.36\pm1.08$        \\
                        ADSSC  &             $91.17\pm6.29$  &             $85.50\pm4.01$  &             $85.83\pm3.49$  &             $82.58\pm2.25$  &             $81.10\pm1.34$  &                $80.03\pm0.79$        \\
  \hline
  \end{tabular}
  \end{center}
  \end{table*}

  \begin{table*}
    \scriptsize
    \begin{center}
    \caption{Segmentation accuracies ($\%$) $\pm$ standard variances of evaluated algorithms on the sub-databases of AR database. The best results (mean) of the algorithms are emphasized in bold.}\label{t6}
    \begin{tabular}{c|ccccc}
    \hline
       \multirow{2}{*}{Methods}  &                     \multicolumn{5}{c}{Number of persons}          \\\cline{2-6}
                                 &                        $4$  &                        $8$  &                       $12$  &                       $16$  &                          $20$       \\\hline
                          IDR-Z  &             $90.29\pm9.44$  &             $88.03\pm7.90$  &             $88.40\pm4.79$  &             $89.59\pm2.38$  &                $87.65\pm4.16$        \\
                          IDR-S  &             $\mathbf{91.44\pm8.45}$  &             $\mathbf{94.18\pm4.67}$  &             $\mathbf{94.04\pm3.89}$  &             $\mathbf{93.08\pm3.22}$  &                $\mathbf{90.54\pm3.53}$        \\\hline
                            SSC  &            $88.27\pm11.28$  &             $82.69\pm9.03$  &             $81.92\pm6.63$  &             $79.18\pm4.04$  &                $79.25\pm4.33$        \\\hline
                            LRR  &            $86.63\pm10.22$  &             $85.53\pm6.89$  &             $87.98\pm6.12$  &             $87.28\pm3.63$  &                $86.27\pm3.04$        \\\hline
                            LSR  &            $83.65\pm12.22$  &             $85.10\pm7.31$  &             $89.78\pm4.66$  &             $86.56\pm3.25$  &                $86.04\pm3.98$        \\\hline
                          BDR-Z  &            $85.29\pm13.40$  &             $83.51\pm6.03$  &             $85.83\pm6.43$  &             $86.13\pm3.94$  &                $83.35\pm2.77$        \\
                          BDR-B  &            $88.17\pm11.00$  &             $85.67\pm7.28$  &             $84.65\pm5.32$  &             $82.74\pm3.77$  &                $81.58\pm4.32$        \\\hline
                         MR-SSC  &            $88.17\pm13.58$  &             $84.23\pm7.93$  &             $82.95\pm7.64$  &             $82.12\pm4.20$  &                $79.71\pm3.47$        \\
                         MR-LRR  &            $84.71\pm14.03$  &             $81.15\pm4.57$  &             $86.25\pm7.23$  &             $88.51\pm3.67$  &                $86.11\pm6.68$        \\\hline
                          JDSSC  &            $66.83\pm19.56$  &             $68.87\pm7.67$  &             $76.60\pm6.83$  &             $80.65\pm2.13$  &                $76.73\pm4.30$        \\
                          ADSSC  &            $83.41\pm12.86$  &             $80.17\pm5.72$  &             $83.41\pm1.72$  &             $83.53\pm1.67$  &                $81.49\pm4.03$        \\
    \hline
  \end{tabular}
  \end{center}
\end{table*}
Clearly, the two tables show that on most cases, IDR outperforms other algorithms on the two databases. Especially on AR database, IDR gets much better results than those of other evaluated algorithms.   
\par We also compare the computation time of all the evaluated algorithms. For a face images database, on its sub-databases with a fixed $q$ (number of persons), we could compute the average computation time of each algorithm. Then the computation time of each algorithm changed with $q$ could be illustrated in the following Fig. \ref{f8}. Similar to the results obtained on Hopkins 155 databases, it could be seen that the computation time of IDR is acceptable. When $q$ is relatively small, the computation cost of IDR is close to that of LRR, when  $q$ is relatively large, IDR is more efficient than LRR. However, JDSSC spends much more time than other algorithms.
\begin{figure}
  \centering
  \includegraphics[width=0.5\textwidth]{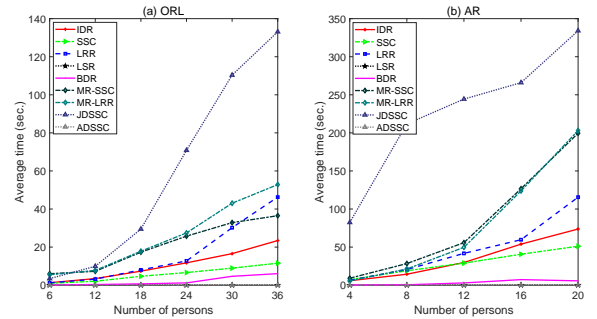}\\
  \caption{The average computation time (sec.) of each algorithm versus the number of persons. (a) the results obtained on ORL database, (b) the results obtained on AR database. }\label{f8}
\end{figure}
\par Finally, we test the convergence of IDR by using all the samples in ORL database. The residuals defined as $\mathbf{redisualZ} = \|\mathbf{Z}^h-\mathbf{Z}^{h+1}\|_F^2,\mathbf{residualS} = \|\mathbf{S}^h-\mathbf{S}^{h+1}\|_F^2,\mathbf{residualE} = \|\mathbf{E}^h-\mathbf{E}^{h+1}\|_F^2$ of the three variables $\mathbf{Z},\mathbf{S},\mathbf{E}$ in Eq. (\ref{e9}). Fig. \ref{e9} plots the residuals versus the number of iterations. It can be seen that the variables $\mathbf{Z},\mathbf{S},\mathbf{E}$ could converge to a station point with a relative small number of iterations by using \textbf{Algorithm 1}. And when the number of iterations is larger than $200$, the residuals are closed to $0$.
\begin{figure*}
  \centering
  \includegraphics[width=0.8\textwidth]{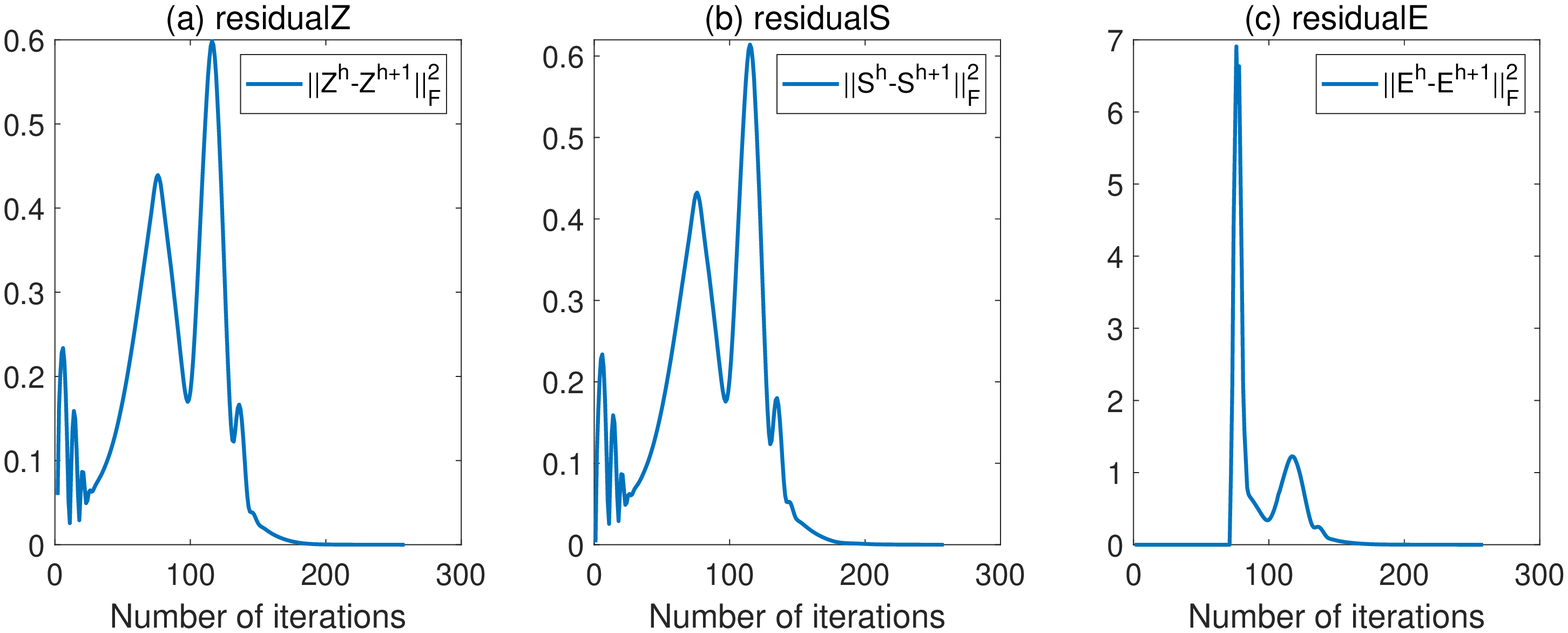}\\
  \caption{The residuals of variables $\mathbf{Z},\mathbf{S},\mathbf{E}$ versus the iterations on the whole ORL database.}\label{f9}
\end{figure*}
\par
\begin{figure}
  \centering
  \includegraphics[width=0.5\textwidth]{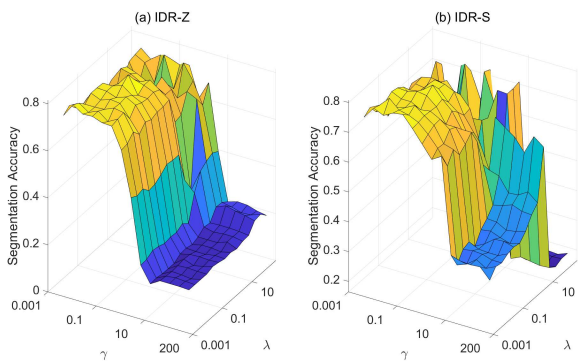}\\
  \caption{The segmentation accuracies of IDR against the variation of parameters on the whole ORL database.}\label{f10}
\end{figure}

 The performances of the evaluated algorithms on the whole ORL database are reported in Table \ref{t7}. We can see that IDR still achieves best results. In Table \ref{t7}, the average computation time of each algorithm with different parameters is also reported. Moreover, the sensitivity verification of IDR to its parameters is illustrated in Fig. \ref{f10}. It still shows that IDR is stable and can get good results when $\gamma$ is relatively small.
\begin{table*}
  \scriptsize
  \begin{center}
  \caption{The segmentation accuracies ($\%$) and average computation time (sec.) of the evaluated algorithms on the whole ORL database. The best results  of the algorithms are emphasized in bold.}\label{t7}
  \begin{tabular}{c|cc|c|c|c|cc|c|c|c|c}
    \hline
   Methods&IDR-Z  & IDR-S &  SSC & LRR & LSR & BDR-Z & BDR-B  & MR-SSC  & MR-LRR & JDSSC  & ADSSC \\\hline
   Segmentation Accuracy&$\mathbf{81.50}$  & $80.75$  & $78.00$ & $76.00$ & $78.25$ & $79.50$ & $81.25$ & $80.25$ & $75.25$ & $73.75$ & $78.25$ \\
    Average time (sec.)&\multicolumn{2}{c|}{$36.19$}  & $16.23$ & $58.55$  & $\mathbf{0.06}$  & \multicolumn{2}{c|}{$26.54$} & $113.84$  & $136.78$  & $88.43$  & $0.11$\\\hline
  \end{tabular}
  \end{center}
\end{table*}

\subsection{Experiments on MNIST data set} 
\par
\begin{figure}
  \centering
  \includegraphics[width=0.45\textwidth]{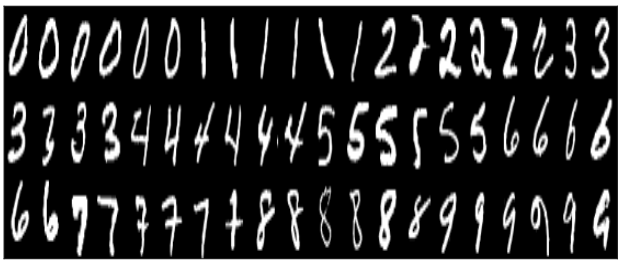}\\
  \caption{Sample images from MNIST database.}\label{f11}
\end{figure}

 MNIST database has 10 subjects, corresponding to $10$ handwritten digits, namely `$0$'-`$9$'. We first select a subset which consists of the first 100 samples of each subject’s training data set to form a sub MNIST database. And each image is resized to $28\times28$ pixels. Some sample images from the database are illustrated in Fig. \ref{f11}. 
\par Then we followed the similar methodologies used in the above experiments. Here, we randomly chose images of $\{2,4,6,8,10\}$ digits from each subjects' training data set to build sub-databases. We also run the experiments $10$ trails and record the mean and standard variance of segmentation accuracies obtained by each algorithm in Table \ref{t8}. 
\begin{table*}
  \scriptsize
  \begin{center}
  \caption{Segmentation accuracies ($\%$) $\pm$ standard variances of evaluated algorithms on the sub-databases of MNIST database. The best results (mean) of the algorithms are emphasized in bold.}\label{t8}
  \begin{tabular}{c|cccccc}
  \hline
     \multirow{2}{*}{Methods}  &                     \multicolumn{5}{c}{Number of digits}          \\\cline{2-6}
                               &                        $2$  &                        $4$  &                        $6$  &                        $8$  &                          $10$       \\\hline
                        IDR-Z  &             $98.45\pm1.99$  &             $84.92\pm7.37$  &             $73.67\pm4.25$  &             $\mathbf{71.26\pm4.13}$  &                $66.65\pm2.38$        \\
                        IDR-S  &             $\mathbf{99.35\pm0.91}$  &             $\mathbf{85.60\pm9.47}$  &             $\mathbf{74.58\pm4.03}$  &             $70.97\pm4.35$  &                $\mathbf{68.00\pm2.09}$        \\ \hline
                          SSC  &             $97.05\pm2.03$  &            $77.15\pm10.67$  &             $67.83\pm3.48$  &             $64.56\pm4.08$  &                $62.71\pm2.15$        \\\hline
                          LRR  &             $96.80\pm2.54$  &             $82.50\pm7.36$  &             $68.38\pm3.70$  &             $64.46\pm4.10$  &                $61.02\pm2.09$        \\\hline
                          LSR  &             $94.70\pm5.42$  &             $77.05\pm7.54$  &             $66.73\pm4.67$  &             $61.29\pm5.50$  &                $56.55\pm2.53$        \\\hline
                          BDR-Z  &             $95.15\pm5.29$  &             $76.35\pm6.55$  &             $67.87\pm4.85$  &             $60.78\pm3.41$  &                $57.79\pm2.11$        \\
                          BDR-B  &             $93.20\pm6.33$  &             $74.78\pm7.56$  &             $64.72\pm6.83$  &             $56.92\pm3.47$  &                $54.21\pm1.65$        \\\hline
                       MR-SSC  &             $97.35\pm1.73$  &             $76.72\pm9.17$  &             $67.33\pm2.91$  &             $63.02\pm2.66$  &                   $59.54\pm1.65$        \\
                       MR-LRR  &             $96.25\pm2.76$  &             $77.40\pm9.33$  &             $66.25\pm3.27$  &             $61.46\pm3.51$  &                   $57.67\pm2.98$        \\\hline
                        JDSSC  &             $97.75\pm1.57$  &            $77.38\pm13.53$  &             $68.73\pm5.14$  &             $63.96\pm4.33$  &                $62.15\pm1.75$        \\
                        ADSSC  &             $95.15\pm4.74$  &             $76.13\pm6.96$  &             $65.33\pm4.78$  &             $60.24\pm3.93$  &                $56.87\pm2.79$        \\
  \hline
  \end{tabular}
  \end{center}
  \end{table*}
\par Form Table \ref{t8}, we could find that IDR still dominates the other algorithms. Actually, IDR achieves much better results than those of other algorithms. In addition, we could see that the performances of other algorithms are closed to each other.
\par
\begin{figure*}
  \centering
  \includegraphics[width=0.8\textwidth]{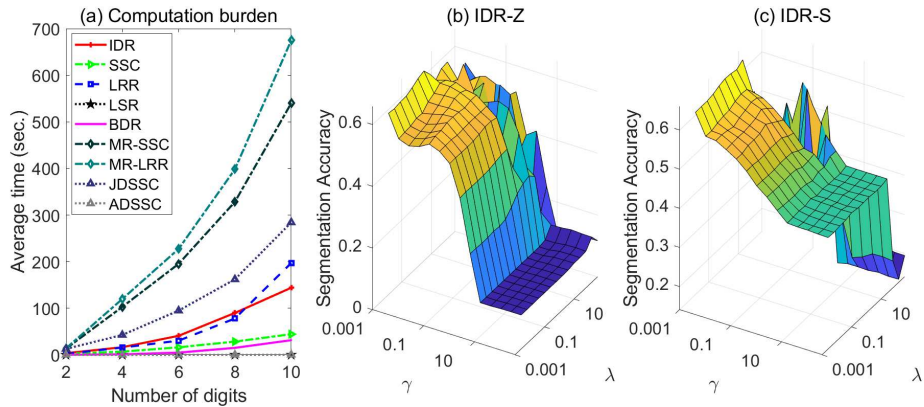}\\
  \caption{(a) The average computation time (sec.) of each algorithm versus the number of digits. (b) The segmentation accuracies of IDR-Z against the variation of parameters. (b) The segmentation accuracies of IDR-S against the variation of parameters. }\label{f12}
\end{figure*}

 Moreover, we also plot the average computation time of each algorithm against the number of digits in Fig. \ref{f12}(a) and show that the performances of IDR-Z and IDR-S changed with the values of parameters $\gamma$ and $\lambda$ in Fig. \ref{f12}(b) and Fig. \ref{f12}(c) respectively. For the visualization of IDR's sensitivity, here we use $10$ sub-databases with $10$ digits. 
\par From Fig. \ref{f12}, we could conclude that 1) the computation time of IDR is much less than MR and JDSSC; 2) the computation costs of MR-SSC and MR-LRR are much larger than those of other algorithms; 2) IDR could achieve better results with small $\gamma$ and $\lambda$. This coincides with the experiments provided above.

\par Based on all the above experiments, we could make the following summarizations: 1) IDR could get satisfying subspace clustering results on different kinds of databases; 2) Compared with the closed related algorithms, such as MR and DSSC, the computation cost of IDR is acceptable; 3) IDR is insensitive to its two parameters. However, small parameters could make IDR achieve better results. 

\section{Conclusions}
\label{sec6}
Spectral-type subspace clustering algorithms show their excellent performances in subspace clustering tasks. The classical spectral-type methods hope to use different norms of reconstruction coefficient matrices to seek coefficient matrices satisfying intra-subspace connectivity and inter-subspace sparse. In this paper, we design an idempotent constraint for reconstruction coefficient matrices based on the proposition that reconstruction coefficient vectors also obey the self-expressiveness property. By integrating double stochastic constraints, we present an idempotent representation (IDR) method for subspace clustering. subspace clustering experiments conducted on both synthetic data sets and real world data sets verify the effectiveness and efficiency of IDR.
\ifCLASSOPTIONcaptionsoff
  \newpage
\fi



%
\bibliographystyle{IEEEtran}
\bibliography{ref}

\begin{thebibliography}{10}
\providecommand{\url}[1]{#1}
\csname url@samestyle\endcsname
\providecommand{\newblock}{\relax}
\providecommand{\bibinfo}[2]{#2}
\providecommand{\BIBentrySTDinterwordspacing}{\spaceskip=0pt\relax}
\providecommand{\BIBentryALTinterwordstretchfactor}{4}
\providecommand{\BIBentryALTinterwordspacing}{\spaceskip=\fontdimen2\font plus
\BIBentryALTinterwordstretchfactor\fontdimen3\font minus
  \fontdimen4\font\relax}
\providecommand{\BIBforeignlanguage}[2]{{%
\expandafter\ifx\csname l@#1\endcsname\relax
\typeout{** WARNING: IEEEtran.bst: No hyphenation pattern has been}%
\typeout{** loaded for the language `#1'. Using the pattern for}%
\typeout{** the default language instead.}%
\else
\language=\csname l@#1\endcsname
\fi
#2}}
\providecommand{\BIBdecl}{\relax}
\BIBdecl

\bibitem{art_1}
L.~Parsons, E.~Haque, and H.~Liu, ``Subspace clustering for high dimensional
  data: A review,'' \emph{SIGKDD Explor. Newsl.}, vol.~6, no.~1, pp. 90--105,
  2004.

\bibitem{RN2518}
R.~Vidal, ``Subspace clustering,'' \emph{IEEE Signal Processing Magazine},
  vol.~28, no.~2, pp. 52--68, 2011.

\bibitem{DBLP:conf/sigmod/AgrawalGGR98}
R.~Agrawal, J.~Gehrke, D.~Gunopulos, and P.~Raghavan, ``Automatic subspace
  clustering of high dimensional data for data mining applications,'' in
  \emph{{SIGMOD} 1998, Proceedings {ACM} {SIGMOD} International Conference on
  Management of Data, June 2-4, 1998, Seattle, Washington, {USA}}, L.~M. Haas
  and A.~Tiwary, Eds.\hskip 1em plus 0.5em minus 0.4em\relax {ACM} Press, 1998,
  pp. 94--105.

\bibitem{DBLP:journals/tkdd/KriegelKZ09}
H.~Kriegel, P.~Kr{\"{o}}ger, and A.~Zimek, ``Clustering high-dimensional data:
  {A} survey on subspace clustering, pattern-based clustering, and correlation
  clustering,'' \emph{{ACM} Trans. Knowl. Discov. Data}, vol.~3, no.~1, pp.
  1:1--1:58, 2009.

\bibitem{RN1940}
R.~Vidal and P.~Favaro, ``Low rank subspace clustering (lrsc),'' \emph{Pattern
  Recognition Letters}, vol.~43, pp. 47--61, 2014.

\bibitem{DBLP:journals/tkde/ChuCYC09}
Y.~Chu, Y.~Chen, D.~Yang, and M.~Chen, ``Reducing redundancy in subspace
  clustering,'' \emph{{IEEE} Trans. Knowl. Data Eng.}, vol.~21, no.~10, pp.
  1432--1446, 2009.

\bibitem{DBLP:journals/tcsv/YiHCC18}
S.~Yi, Z.~He, Y.~Cheung, and W.~Chen, ``Unified sparse subspace learning via
  self-contained regression,'' \emph{{IEEE} Trans. Circuits Syst. Video
  Technol.}, vol.~28, no.~10, pp. 2537--2550, 2018.

\bibitem{DBLP:journals/pami/MaDHW07}
Y.~Ma, H.~Derksen, W.~Hong, and J.~Wright, ``Segmentation of multivariate mixed
  data via lossy data coding and compression,'' \emph{{IEEE} Trans. Pattern
  Anal. Mach. Intell.}, vol.~29, no.~9, pp. 1546--1562, 2007.

\bibitem{DBLP:journals/pami/ElhamifarV13}
E.~Elhamifar and R.~Vidal, ``Sparse subspace clustering: algorithm, theory, and
  applications,'' \emph{IEEE Trans Pattern Anal Mach Intell}, vol.~35, no.~11,
  pp. 2765--81, 2013.

\bibitem{RN1710}
G.~Liu, Z.~Lin, S.~Yan, J.~Sun, Y.~Yu, and Y.~Ma, ``Robust recovery of subspace
  structures by low-rank representation,'' \emph{IEEE Transaction on Pattern
  Analysis and Machine Intelligence}, vol.~35, no.~1, pp. 171--184, 2013.

\bibitem{DBLP:conf/iccv/Kanatani01}
K.~Kanatani, ``Motion segmentation by subspace separation and model
  selection,'' in \emph{Proceedings of the Eighth International Conference On
  Computer Vision (ICCV-01), Vancouver, British Columbia, Canada, July 7-14,
  2001 - Volume 2}.\hskip 1em plus 0.5em minus 0.4em\relax {IEEE} Computer
  Society, 2001, pp. 586--591.

\bibitem{DBLP:journals/siamrev/MaYDF08}
Y.~Ma, A.~Y. Yang, H.~Derksen, and R.~M. Fossum, ``Estimation of subspace
  arrangements with applications in modeling and segmenting mixed data,''
  \emph{{SIAM} Review}, vol.~50, no.~3, pp. 413--458, 2008.

\bibitem{DBLP:journals/pami/ShiM00}
J.~Shi and J.~Malik, ``Normalized cuts and image segmentation,'' \emph{{IEEE}
  Trans. Pattern Anal. Mach. Intell.}, vol.~22, no.~8, pp. 888--905, 2000.

\bibitem{conf_1}
E.~Elhamifar and R.~Vidal, ``Sparse subspace clustering,'' in \emph{Proc.
  {IEEE} Computer Society Conference on Computer Vision and Pattern Recognition
  ({CVPR} 2009)}, Miami, Florida, USA, Jun. 2009, pp. 2790--2797.

\bibitem{RN802}
J.~Wright, A.~Y. Yang, A.~Ganesh, S.~S. Sastry, and Y.~Ma, ``Robust face
  recognition via sparse representation,'' \emph{IEEE Transactions on Pattern
  Analysis and Machine Intelligence}, vol.~31, no.~2, pp. 210--227, 2009.

\bibitem{conf_2}
G.~Liu, Z.~Lin, and Y.~Yu, ``Robust subspace segmentation by low-rank
  representation,'' in \emph{Proceedings of the 27th International Conference
  on Machine Learning (ICML-10), June 21-24, 2010,}, Haifa, Israel, jun 2010,
  pp. 663--670.

\bibitem{Lu:2012}
C.-Y. Lu, H.~Min, Z.-Q. Zhao, L.~Zhu, D.-S. Huang, and S.~Yan, ``Robust and
  efficient subspace segmentation via least squares regression,'' in
  \emph{Proceedings of the 12th European Conference on Computer Vision, {ECCV}
  2012}, Florence, Italy, 2012, pp. 347--360.

\bibitem{RN2485}
C.~Lu, J.~Feng, Z.~Lin, T.~Mei, and S.~Yan, ``Subspace clustering by block
  diagonal representation,'' \emph{IEEE Transactions on Pattern Analysis and
  Machine Intelligence}, vol.~41, no.~2, pp. 487--501, 2019.

\bibitem{RN2478}
H.~Zhang, J.~Yang, F.~Shang, C.~Gong, and Z.~Zhang, ``Lrr for subspace
  segmentation via tractable schatten-p norm minimization and factorization,''
  \emph{IEEE Transactions on Cybernetics}, pp. 1--13, 2018.

\bibitem{RN2691}
J.~Xu, M.~Yu, L.~Shao, W.~Zuo, D.~Meng, L.~Zhang, and D.~Zhang, ``Scaled
  simplex representation for subspace clustering,'' \emph{IEEE TRANSACTIONS ON
  CYBERNETICS}, 2019.

\bibitem{RN2328}
C.-G. Li, C.~You, and R.~Vidal, ``Structured sparse subspace clustering: A
  joint affinity learning and subspace clustering framework,'' \emph{IEEE Trans
  Image Process}, vol.~26, no.~6, pp. 2988--3001, 2017.

\bibitem{RN1705}
Y.~Panagakis and C.~Kotropoulos, ``Elastic net subspace clustering applied to
  pop/rock music structure analysis,'' \emph{Pattern Recognition Letters},
  vol.~38, pp. 46--53, 2014.

\bibitem{RN2562}
C.~You, C.~Li, D.~P. Robinson, and R.~Vidal, ``Oracle based active set
  algorithm for scalable elastic net subspace clustering,'' in \emph{2016
  {IEEE} Conference on Computer Vision and Pattern Recognition, {CVPR} 2016,
  Las Vegas, NV, USA, June 27-30, 2016}.\hskip 1em plus 0.5em minus 0.4em\relax
  {IEEE} Computer Society, 2016, pp. 3928--3937.

\bibitem{DBLP:journals/tip/ZhuangGTWLMY15}
L.~Zhuang, S.~Gao, J.~Tang, J.~Wang, Z.~Lin, Y.~Ma, and N.~Yu, ``Constructing a
  nonnegative low-rank and sparse graph with data-adaptive features,''
  \emph{{IEEE} Trans. Image Processing}, vol.~24, no.~11, pp. 3717--3728, 2015.

\bibitem{RN1888}
K.~Tang, R.~Liu, and J.~Zhang, ``Structure-constrained low-rank
  representation,'' \emph{IEEE TRANSACTIONS ON NEURAL NETWORKS AND LEARNING
  SYSTEMS}, vol.~25, no.~12, pp. 2167--2179, 2014.

\bibitem{RN1946}
X.~Lu, Y.~Wang, and Y.~Yuan, ``Graph-regularized low-rank representation for
  destriping of hyperspectral images,'' \emph{IEEE Transaction on Geoscience
  and Remote Sensing}, vol.~51, no. 7-1, pp. 4009--4018, 2013.

\bibitem{RN2192}
K.~Tang, D.~B. Dunson, Z.~Su, R.~Liu, J.~Zhang, and J.~Dong, ``Subspace
  segmentation by dense block and sparse representation,'' \emph{Neural
  Network}, vol.~75, pp. 66--76, 2016.

\bibitem{RN2723}
L.~Wei, F.~Ji, H.~Liu, R.~Zhou, C.~Zhu, and X.~Zhang, ``Subspace clustering via
  structured sparse relation representation,'' \emph{IEEE Trans Neural Network
  and Learn Systems}, vol.~PP, 2021.

\bibitem{RN2532}
Y.~Sui, G.~Wang, and L.~Zhang, ``Sparse subspace clustering via low-rank
  structure propagation,'' \emph{Pattern Recognition}, vol.~95, pp. 261--271,
  2019.

\bibitem{DBLP:conf/soda/MathieuS10}
C.~Mathieu and W.~Schudy, ``Correlation clustering with noisy input,'' in
  \emph{Proceedings of the Twenty-First Annual {ACM-SIAM} Symposium on Discrete
  Algorithms, {SODA} 2010, Austin, Texas, USA, January 17-19, 2010},
  M.~Charikar, Ed.\hskip 1em plus 0.5em minus 0.4em\relax {SIAM}, 2010, pp.
  712--728.

\bibitem{DBLP:conf/soda/Swamy04}
C.~Swamy, ``Correlation clustering: maximizing agreements via semidefinite
  programming,'' in \emph{Proceedings of the Fifteenth Annual {ACM-SIAM}
  Symposium on Discrete Algorithms, {SODA} 2004, New Orleans, Louisiana, USA,
  January 11-14, 2004}, J.~I. Munro, Ed.\hskip 1em plus 0.5em minus 0.4em\relax
  {SIAM}, 2004, pp. 526--527.

\bibitem{DBLP:conf/cvpr/LeeLLK15}
M.~Lee, J.~Lee, H.~Lee, and N.~Kwak, ``Membership representation for detecting
  block-diagonal structure in low-rank or sparse subspace clustering,'' in
  \emph{{IEEE} Conference on Computer Vision and Pattern Recognition, {CVPR}
  2015, Boston, MA, USA, June 7-12, 2015}.\hskip 1em plus 0.5em minus
  0.4em\relax {IEEE} Computer Society, 2015, pp. 1648--1656.

\bibitem{DBLP:journals/corr/LinCM10}
Z.~Lin, M.~Chen, and Y.~Ma, ``The augmented lagrange multiplier method for
  exact recovery of corrupted low-rank matrices,'' \emph{CoRR}, vol.
  abs/1009.5055, 2010.

\bibitem{DBLP:conf/iccv/ZassS05}
R.~Zass and A.~Shashua, ``A unifying approach to hard and probabilistic
  clustering,'' in \emph{10th {IEEE} International Conference on Computer
  Vision {(ICCV} 2005), 17-20 October 2005, Beijing, China}.\hskip 1em plus
  0.5em minus 0.4em\relax {IEEE} Computer Society, 2005, pp. 294--301.

\bibitem{DBLP:conf/nips/ZassS06a}
------, ``Doubly stochastic normalization for spectral clustering,'' in
  \emph{Advances in Neural Information Processing Systems 19, Proceedings of
  the Twentieth Annual Conference on Neural Information Processing Systems,
  Vancouver, British Columbia, Canada, December 4-7, 2006}, B.~Sch{\"{o}}lkopf,
  J.~C. Platt, and T.~Hofmann, Eds.\hskip 1em plus 0.5em minus 0.4em\relax
  {MIT} Press, 2006, pp. 1569--1576.

\bibitem{DBLP:books/lib/DudaHS01}
R.~O. Duda, P.~E. Hart, and D.~G. Stork, \emph{Pattern classification, 2nd
  Edition}.\hskip 1em plus 0.5em minus 0.4em\relax Wiley, 2001.

\bibitem{DBLP:conf/icml/Mairal13}
J.~Mairal, ``Optimization with first-order surrogate functions,'' in
  \emph{Proceedings of the 30th International Conference on Machine Learning,
  {ICML} 2013, Atlanta, GA, USA, 16-21 June 2013}, ser. {JMLR} Workshop and
  Conference Proceedings, vol.~28.\hskip 1em plus 0.5em minus 0.4em\relax
  JMLR.org, 2013, pp. 783--791.

\bibitem{DBLP:journals/corr/abs-2011-14859}
D.~Lim, R.~Vidal, and B.~D. Haeffele, ``Doubly stochastic subspace
  clustering,'' \emph{CoRR}, vol. abs/2011.14859, 2020.

\bibitem{DBLP:journals/jscic/Ma16}
S.~Ma, ``Alternating proximal gradient method for convex minimization,''
  \emph{J. Sci. Comput.}, vol.~68, no.~2, pp. 546--572, 2016.

\bibitem{RN2445}
L.~Wei, X.~Wang, A.~Wu, R.~Zhou, and C.~Zhu, ``Robust subspace segmentation by
  self-representation constrained low-rank representation,'' \emph{Neural
  Processing Letters}, vol.~48, no.~3, pp. 1671--1691, 2018.

\bibitem{RN2572}
C.~You, C.-G. Li, D.~P. Robinson, and R.~Vidal, ``Is an affine constraint
  needed for affine subspace clustering?'' \emph{ICCV}, 2019.

\bibitem{DBLP:conf/cvpr/TronV07}
R.~Tron and R.~Vidal, ``A benchmark for the comparison of 3-d motion
  segmentation algorithms,'' in \emph{2007 {IEEE} Computer Society Conference
  on Computer Vision and Pattern Recognition {(CVPR} 2007), 18-23 June 2007,
  Minneapolis, Minnesota, {USA}}.\hskip 1em plus 0.5em minus 0.4em\relax {IEEE}
  Computer Society, 2007.

\bibitem{DBLP:conf/wacv/SamariaH94}
F.~Samaria and A.~Harter, ``Parameterisation of a stochastic model for human
  face identification,'' in \emph{Proceedings of Second {IEEE} Workshop on
  Applications of Computer Vision, {WACV} 1994, Sarasota, FL, USA, December
  5-7, 1994}.\hskip 1em plus 0.5em minus 0.4em\relax {IEEE}, 1994, pp.
  138--142.

\bibitem{ARface}
A.~Martinez and R.~Benavente, ``The ar face database,'' \emph{CVC Technical
  Report 24}, 1998.

\end{thebibliography}

\end{document}